\documentclass{article}

\usepackage[nonatbib,final]{neurips_2025}

\usepackage{microtype}
\usepackage{graphicx}
\usepackage{subfigure}
\usepackage{booktabs} %

\usepackage{hyperref}

\usepackage{amsmath}
\usepackage{amssymb}
\usepackage{mathtools}
\usepackage{amsthm}
\usepackage[table]{xcolor}
\usepackage{multirow}
\usepackage{caption}

\usepackage[capitalize,noabbrev]{cleveref}

\theoremstyle{plain}

\theoremstyle{definition}

\theoremstyle{remark}

\usepackage[textsize=tiny]{todonotes}

\definecolor{third}{rgb}{1, 1, 0.7}
\definecolor{second}{rgb}{1, 0.85, 0.7}
\definecolor{first}{rgb}{1, 0.7, 0.7}
\definecolor{lightblue}{rgb}{0.21, 0.8, 0.92}
\definecolor{backprop}{rgb}{0.5, 0.0, 0.5}

\newcommand{\cfirst}{\cellcolor{lightblue}}

\definecolor{progressive}{rgb}{1, 0.4, 0.0}
\definecolor{qat}{rgb}{0, 0.5, 0.0}
\definecolor{discrete}{rgb}{0, 0, 1.0}

\newcommand{\secref}[1]{Section \ref{#1}}
\newcommand{\figref}[1]{Figure \ref{#1}}
\newcommand{\tabref}[1]{Table \ref{#1}}

\newcommand{\eg}{e.g.}
\newcommand{\ie}{i.e.}
\newcommand{\etal}{et al.}
\newcommand{\rows}[2]{\multirow{#1}{*}{#2}}
\newcommand{\cols}[2]{\multicolumn{#1}{c}{#2}}

\newcommand{\name}{Fractional Quantization Aware Training}
\newcommand{\short}{FraQAT}

\newcommand{\pixart}{PixArt-$\Sigma$}

\newcommand{\Q}{\mathcal{Q}}
\newcommand{\Real}{\mathbb{R}}
\newcommand{\W}{\boldsymbol{W}}
\newcommand{\T}{\mathbb{T}}

\newcommand{\FP}[1]{\texttt{FP#1}}
\newcommand{\INT}[1]{\texttt{INT#1}}
\newcommand{\WA}[2]{\texttt{W#1A#2}}

\title{FraQAT: Quantization Aware Training with Fractional bits}

\usepackage{ulem,algorithm,algpseudocode}
\algnewcommand\algorithmicinput{\textbf{Input:}}
\algnewcommand\algorithmicoutput{\textbf{Output:}}
\algnewcommand\Input{\item[\algorithmicinput]}%
\algnewcommand\Output{\item[\algorithmicoutput]}%

\author{%
  Luca Morreale\\
  \And
  Alberto Gil C. P. Ramos\\
  \And
  Malcolm Chadwick\\
  \And
  Mehid Noroozi\\
  \And
  Ruchika Chavhan\\
  \And
  Abhinav Mehrotra\\ \\
  Samsung AI Center\\
  Cambridge, UK\\
  \And
  Sourav Bhattacharya\\
}

\begin{document}

\maketitle

\begin{figure}[htbp]
    \centering
    \setlength{\tabcolsep}{0.0pt}
    \begin{tabular}{cc@{\hskip 2pt}cc@{\hskip 2pt}cc}
    	\includegraphics[width=0.166\linewidth]{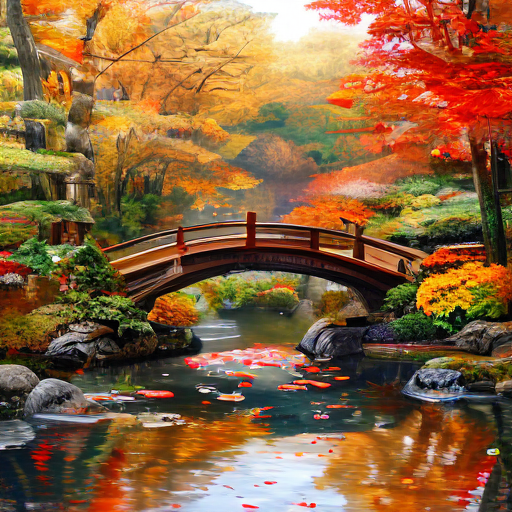} &
    	\includegraphics[width=0.166\linewidth]{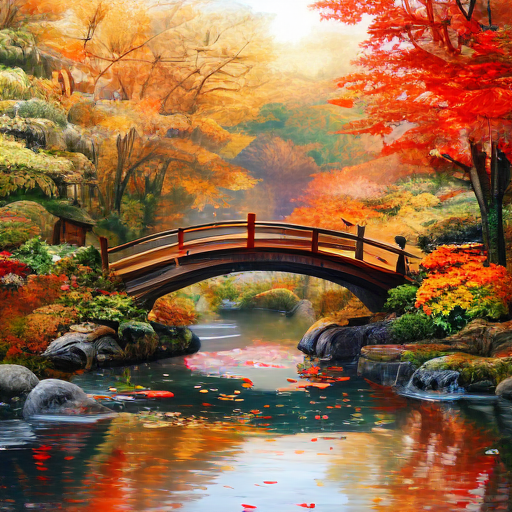} &
    	\includegraphics[width=0.166\linewidth]{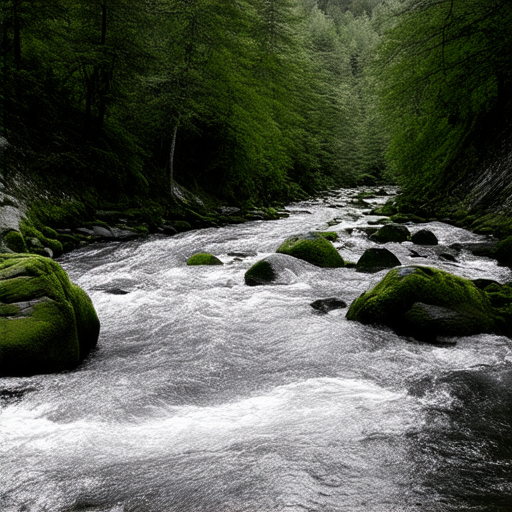} &
    	\includegraphics[width=0.166\linewidth]{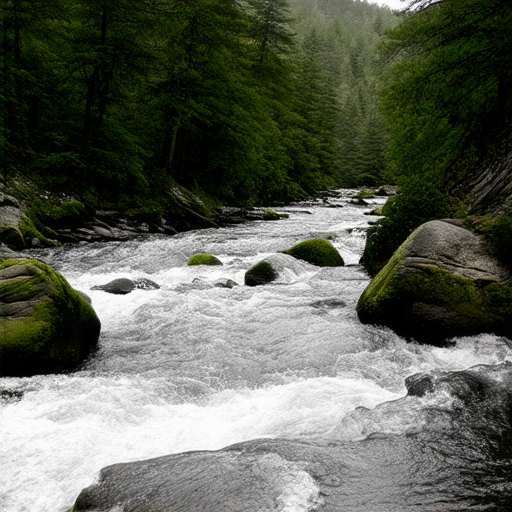} &
    	\includegraphics[width=0.166\linewidth]{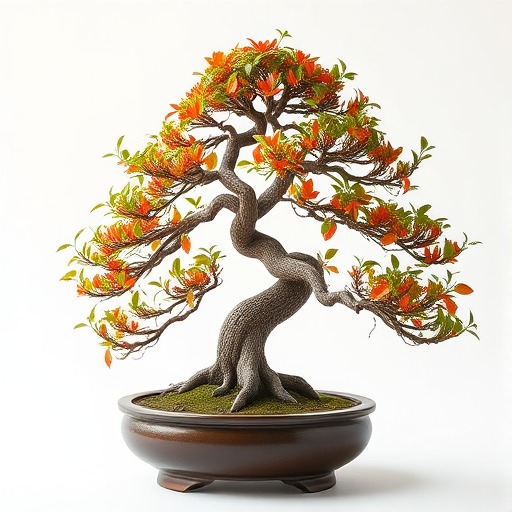} &
    	\includegraphics[width=0.166\linewidth]{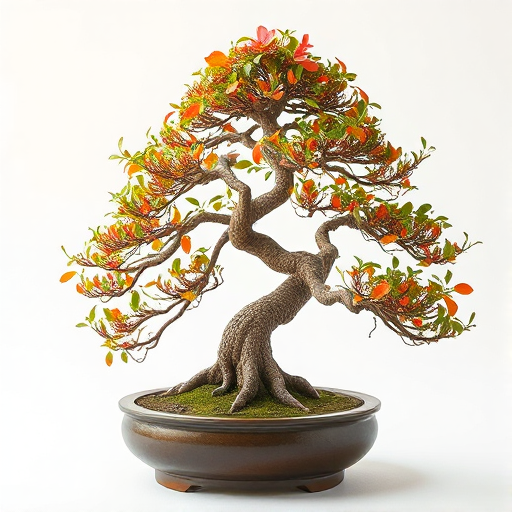} \\
    	Sana 600M & \short & SD3.5-M & \short & Flux-schnell & \short\\
    \end{tabular}
    \caption{\label{fig:teaser}\short~is a Quantization aware Training (QAT) technique that grants generative models high fidelity at a fraction of training time required. Large text-to-image (T2I) models quantized with \short~(\WA{4}{8}) achieve $16\%$ lower FiD score than the state-of-the-art. }
\end{figure}

\begin{abstract}

State-of-the-art (SOTA) generative models have demonstrated impressive capabilities in image synthesis or text generation, often with a large capacity model. However, these large models cannot be deployed on smartphones due to the limited availability of on-board memory and computations. Quantization methods lower the precision of the model parameters, allowing for efficient computations, \eg, in \INT{8}. Although aggressive quantization addresses efficiency and memory constraints, preserving the quality of the model remains a challenge. To retain quality in previous aggressive quantization, we propose a new fractional bits quantization (\short) approach. The novelty is a simple yet effective idea: we progressively reduce the model's precision from 32 to 4 bits per parameter, and exploit the fractional bits during optimization to maintain high generation quality. We show that the \short{} yields improved quality on a variety of diffusion models, including SD3.5-Medium, Sana, \pixart, and FLUX.1-schnell, while achieving $4-7\%$ lower FiD than standard QAT. Finally, we deploy and run Sana on a Samsung S25U, which runs on the Qualcomm SM8750-AB Snapdragon 8 Elite Hexagon Tensor Processor (HTP).

\end{abstract}

\section{Introduction}

Over the past few years, generative models have made impressive progress in synthesizing high-quality images~\cite{sd352024,sana_15,flux2024} and texts~\cite{grattafiori2024llama,team2025gemma}.
Such a breakthrough is partly achieved by enlarging the model's size, \eg, Diffusion Transformer (DiT) models with over 10 billion (10B) parameters are increasingly common.
However, larger models require significantly more resources, hence higher inference-time or latency, even for inference. This increase is particularly problematic for deploying these models on resource limited devices, \eg, smartphones, thus limiting their wide-scale usage.

A well-established approach to mitigate these resource constraints is \textit{quantization}: by shifting parameters from 32 bits to a lower precision, \eg, 4 bits, the model's weight-allocation footprints in its computational graph is significantly reduced. %
While past quantization research aimed mostly at decreasing model size, low-precision hardware support, such as NPUs on smartphones, drives researchers to further decrease inference latency. %
For example, latency gains from reduced data movement are boosted by native support for low-precision operations, \eg, using 4-bit weights and 8-bit activations (\WA{4}{8}). Although, initially few devices offered support for these operations, modern hardware manufacturers readily offer low precision operations across devices: \WA{4}{8} in Qualcomm Snapdragon HTP~\cite{qualcomm_snapdragon}, \INT{8}, \texttt{BF16} and \FP{16} in Intel CPUs~\cite{intel_xeon}, Block \FP{16} in AMD CPUs~\cite{amd_ryzen} and \FP{8}/\FP{4} in NVIDIA GPU H100/H200~\cite{nvidia_hgx} to name a few.

The advantages of deploying cloud-quality generative models on-device are multi-fold: it preserves users' privacy while offering a low-latency experience. For service providers, it reduces operating costs by pushing compute from expensive servers to users' personal devices as well as avoiding violating country-specific privacy regulations. In this work, we target mobile deployment, and restrict ourselves to \WA{4}{8} given its ubiquitous availability across devices.

Quantization approaches fall under two main categories: Post Training Quantization (PTQ) and Quantization-Aware-Training (QAT).
PTQ creates a low-precision model from a high precision pre-trained model using a small calibration dataset. Recent progress in PTQ research has resulted in \WA{4}{32}\footnote{\url{https://github.com/NVlabs/Sana/blob/main/asset/docs/4bit_sana.md}} and \WA{8}{32}\footnote{\url{https://github.com/NVlabs/Sana/blob/main/asset/docs/8bit_sana.md}} high quality quantized models from pre-trained SANA~\cite{sana}, SANA 1.5~\cite{sana_15} and SANA-Sprint~\cite{sana_sprint}. Mixed precision \WA{4}{32} and \WA{16}{32} approaches like SVDQuant~\cite{li2024svdquant} have also yielded high quality quantized models from pre-trained FLUX.1-schnell. In essence, PTQ is ideal for cases where access to a large training dataset or compute cluster is limited.

Despite its success, PTQ requires careful data selection~\cite{zhang2025selectq}. For example, a poorly selected calibration dataset may manifest in poor prompt adherence or exhibit color shifts during deployment. Instead, Quantization-Aware-Training (QAT) optimizes weights in lower precision to boost the overall model's performance~\cite{nair2025matryoshka,bulat2021bit,sui2024bitsfusion}. In general, QAT approaches yield better results, at lower precision, when a large training dataset or a compute cluster is available.
Nonetheless, quantized models suffer from a quality loss compared to the original \FP{32} model.

We propose fractional bits quantization (\short) to bridge the quality gap between the original and the quantized model.
Inspired by Curriculum Learning~\cite{bengio2009curriculum}, our training process progressively increases the quantization complexity, \ie, gradually lowers parameter precision, while replicating the original model's output.
We show that \short{} reduces outliers, stabilizes training and yields improved prompt adherence and image generation quality (\secref{sec:method}). We apply \short{} to the linear layers of SOTA generative models as they contain the majority of the parameters, and empirically demonstrate the advantages of the proposed techniques on diffusion models (\secref{sec:quantitative},\ref{sec:qualitative}). In terms of image quality, \short{} achieves $16\%$ lower FiD than SOTA QAT. To address computational costs, we perform an outlier analysis (\secref{sec:outliers}), and selectively train a subset of the model's layers and show its performance. Finally, we quantize and deploy a model on a Samsung S25U, running on Qualcomm SM8750-AB Snapdragon 8 Elite Hexagon Tensor Processor (HTP) (\secref{sec:on_device}).

\section{Method}

\subsection{Quantization preliminaries} \label{sec:preliminaries}

The goal of quantization is to approximate -- in dynamic or static finite precision -- internal model operations, such as operations within linear layers $\boldsymbol{x} \times \W$ where $\boldsymbol{x} \in \Real^{B\times m}$ and $\W \in \Real^{m\times n}$. Depending on hardware support, the quantization operation $\W_b$ on a matrix $\W$ to $b$ bits can be expressed with narrower range as:

\begin{align}
    \Q(\W)_{b}
    &\coloneqq
    \left\lfloor
    \frac{2^{b - 1} - 1}{\max_{i,j}|[\W]_{i,j}|}
    \W
    \right\rfloor \, \in\{-2^{b-1},\ldots,2^{b-1}-1\}\nonumber\\
    \mathcal{S}(\W)_{b}
    &\coloneqq
    \frac{\max_{i,j}|[\W]_{i,j}|}{2^{b - 1} - 1}
    \in\mathbb{R}^+\nonumber\\
    \W_b
    &\coloneqq
    \mathcal{S}(\W)_{b}\Q(\W)_{b}\label{eq:quantization_without_offset}
\end{align}
or with wider range as:
\begin{align}
    \Q(\W)_{b}
    &\coloneqq
    \left\lfloor
    2^b\frac{\W-w_{\min}}{w_{\max}-w_{\min}}
    \right\rfloor \, \in\{0,\ldots,2^b-1\}\nonumber\\
    \mathcal{S}(\W)_{b}
    &\coloneqq
    \frac{w_{\max}-w_{\min}}{2^b}
    \in\mathbb{R}^+\nonumber\\
    \W_b
    &\coloneqq
    \mathcal{S}(\W)_{b}\Q(\W)_{b}+w_{\min}\label{eq:quantization_with_offset}
\end{align}
where $w_{\min}\coloneqq\min_{i,j}[\W]_{i,j}$ and $w_{\max}\coloneqq\max_{i,j}[\W]_{i,j}$. Most simply for \eqref{eq:quantization_without_offset}, matrix multiplications can be rewritten as:
$x_{b_x}\W_{b_{\W}} = (\mathcal{S}(x)_{b_x}\mathcal{S}(\W)_{b_{\W}})(\Q(x)_{b_x}\Q(\W)_{b_{\W}})$
where $b_x$ and $b_{\W}$ may differ. Therefore, matrix multiplication $x_{b_x}\W_{b_{\W}}$ can be reduced to the multiplication of two floats $\mathcal{S}(x)_{b_x}\mathcal{S}(\W)_{b_{\W}}$ and matrix multiplication of two integer matrices $\Q(x)_{b_x}\Q(\W)_{b_{\W}}$. Furthermore, we refer to dynamic quantization when $w_{\min}$ and $w_{\max}$ are computed at runtime, per sample, based on the input. While, in static quantization, $w_{\min}$ and $w_{\max}$ are pre-computed and shared across all samples. Dynamic quantization, especially when applied to activations, allow to robustly handle outliers as each sample range is computed to maximize representability. On the other hand, static quantization are more restrictive, and generate more outliers, thus making the quantization problem strictly harder. Edge devices, such as smartphones, only support static quantization, while GPUs support both.

In contrast, activations are often quantized through a look-up table mapping from a $2^b$ sized partition of the input range into a fixed number of quantized output values, \eg, the previous layer output $x$. In general, weights and activations may be quantized to different precisions, upcasted to the same precision before computation and downcasted after computation.

Hereafter, we make the number of bits in weights and activations explicit with subscripts, \eg~$\boldsymbol{x}_{32}$ refers to a $32$ bit approximation of $\boldsymbol{x}$.

Due to restricted address spaces in most mobile accelerators, it is critical to decrease weights precision aggressively, especially in large vision or language models, \eg, 12 billion parameter models, otherwise these models cannot even be placed on the target devices. However, naively lowering the weight's precision from \FP{32} to \INT{4} causes severe degradation in the generated results. This degradation is exacerbated by lowering activations precision, as required by integer-only accelerators, most often to \INT{8} for reduced generation latency. At a high level, the quality degradation phenomenon is attributed to outliers in both activations and weights due to training. %
The overall challenge of quantization is to approximate the original network's behavior while lowering the precision:
\begin{equation}
    \boldsymbol{x}\W \approx \boldsymbol{x}_{8}\W_{4} .
    \label{eq:linear_qat}
\end{equation}

\subsection{Fractional Quantization-Aware-Training} \label{sec:method}

\begin{figure}[t]
    \centering
    \includegraphics[width=\textwidth]{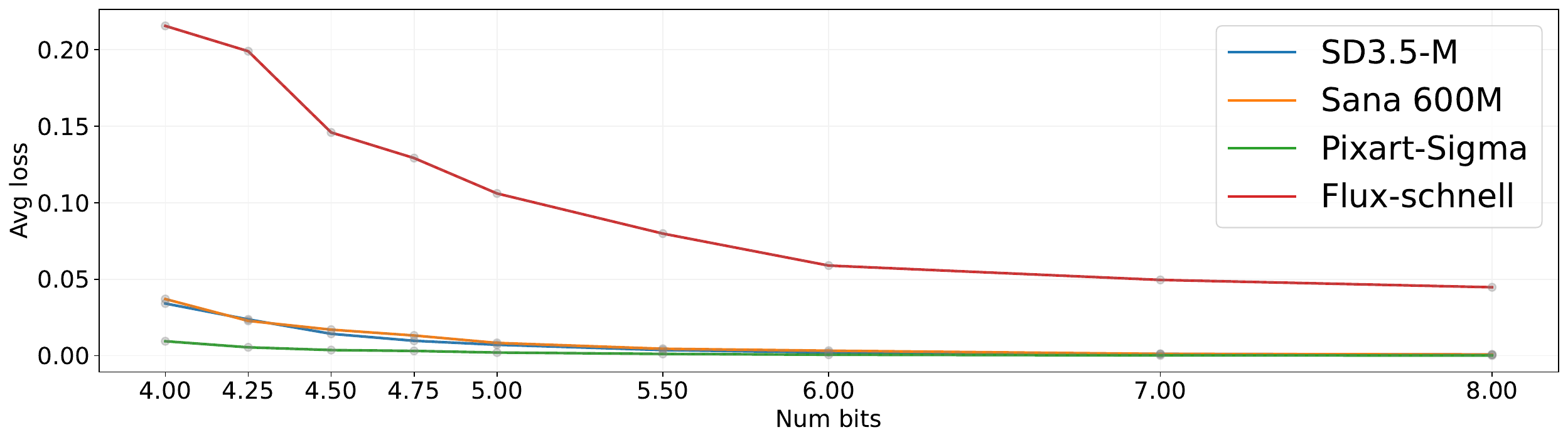} %
    \caption{\textbf{Bit vs Loss:}\label{fig:bitloss}
    as we reduce the precision (number of bits), the average knowledge distillation loss increases -- the gap between student and teacher widens. From a quantization perspective, this implies outliers incrementally affect the student model.}
    \vspace{-0.3cm}
\end{figure}

Intuitively, Quantization-Aware-Training (QAT) approaches -- including the proposed \short{} -- handle outliers, both in weights and activations, by shifting parameters to quantization centroids within or towards adjacent bins. Hence, re-distributing weights in a more compact space. Consequently, the further apart bins are, the harder the optimization problem. We further speculate that it is slower to optimize for lower precisions -- \INT{4} vs \FP{32} -- as the gap between two adjacent representable numbers is much larger. Indeed, it can be observed from \figref{fig:bitloss} that the loss is higher for lower precisions, thus, outliers appear gradually. %

To address this issue, we take inspiration from Curriculum Learning~\cite{bengio2009curriculum} literature: we progressively increase the complexity of the task during the optimization by gradually lowering weights' precision while approximating the full precision model's output. 
This is achieved by two key designs: first \short{} leverages weights from pre-trained models. Second, \short{} continuously steps between discrete quantization ranges to exploit the fact that Eq.~\eqref{eq:quantization_without_offset}--\eqref{eq:quantization_with_offset} are purely a software construct, hence, it is possible to span any continuous --- not just discrete -- precision $b \in [32 ; 1] \in \Real$.

\begin{figure}[t]

\centering
\includegraphics[width=\linewidth]{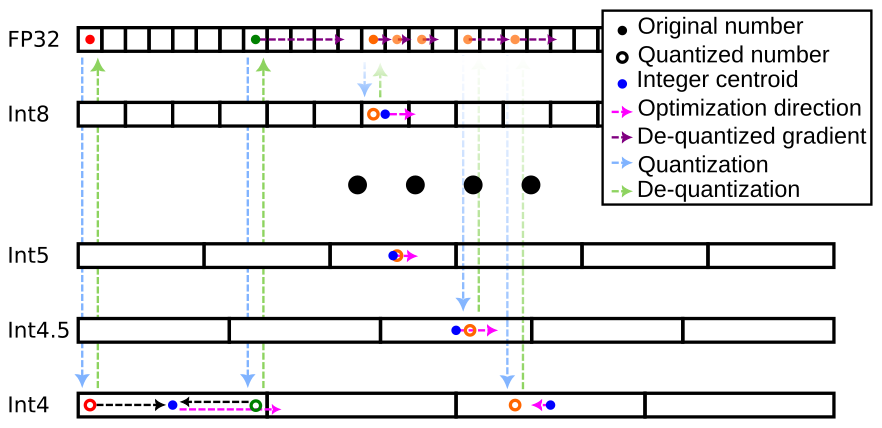}
\caption{\label{fig:fqat}Classic QAT first computes the loss computed at the {\color{qat}lower precision ($\circ$)}, then propagates it back to the original precision and optimize the weights ({\color{backprop}$\rightarrow$}). This results in coarse and noisy gradients. \name{} rely on intermediate precisions {\color{progressive}(from \INT{8} to \INT{4} as training progresses)} to incrementally adjust to weights distributions. Consequently, during training Parameters smoothly shift between {\color{discrete}bins ($\bullet$)} thanks to the finer gradients from in intermediate precision.
}
\end{figure}

Coupled together, these concepts establish \short{} as a faster and higher quality QAT scheme dubbed Fractional Quantization Aware Training (\short). Given a model, \short{} progressively lowers the precision, first coarsely between \FP{32} and \INT{8} and then finely from \INT{8} to \INT{4}, stepping through intermediate fractional bits during training as depicted in \figref{fig:fqat}~and in Algorithm~\ref{alg:algorithm}. As the training progresses, outliers gradually appear, as shown in \figref{fig:bitloss}, and are addressed. Furthermore, by optimizing at fractional bit precision in a curriculum fashion, \short{} allows the weights to move to stable configurations, yielding higher quality samples, and reducing training time. Finally, throughout the entire training process \short{} keeps all activation quantization constant (\INT{8}).

As training progresses, this progressive lowering of precision, smoothly moves weights distribution and thus facilitates quantization
(c.f., Figure \ref{fig:fqat}). Note that it is possible to even set $b=5.5$. Although half-bits precision have no meaning, in practice they bridge the gap of range of representable numbers between two precisions: $\INT{6} \in [-32;31]$, $\INT{5.5} \in [-22;21]$, $\INT{5} \in [-16;15]$. In other words, half-bits precisions reduce the distance between adjacent bins, speeding up convergence without ad-hoc hyper-parameters, such as learning rate.

The proposed Fractional Quantization-Aware-Training (\short) approach is generally applicable to any model and quantization level. Given the wide-spread usage of DiT and MM-DiT blocks in SOTA T2I models, we focus the presentation on DiT models. Since model size limits must first be met for any on-device placement, \short{} quantizes linear layers as they contain the bulk of the parameters of DiT models ($99.9\%$). In particular, \short{} targets the most aggressive \WA{4}{8} quantization, as it allows for a wider range of models to fit edge accelerators with the lowest generation latency. Nonetheless, the proposed technique allows to target any precision.

\begin{algorithm}[b]
    \caption{\name\label{alg:algorithm}}
    \begin{algorithmic}[1]
        \Input Pre-trained model $\mathcal{M}_{W32A32}$, dataset $\mathcal{D}$, loss function $\mathcal{L}$, quantization schedule $\mathcal{B}$ (\eg, $\{8, 5, 4.5, 4\}$), optimizer $\mathcal{O}$
        \Output Quantized model $\mathcal{M}_{W4A8}$
        
        \State $\mathcal{M}_{WbA8} \leftarrow \mathcal{M}_{W32A32}$
        
        \For{$b \in \mathcal{B}$}
        \State $\mathcal{M}_{WbA8} \leftarrow$  \Call{QuantizeLinearLayer}{$\mathcal{M}_{WbA8}$, W$b$A8}
        \For{batch $\in \mathcal{D}$}
        \State $O_{WbA8} \leftarrow$ \Call{forward}{$\mathcal{M}_{WbA8}$, batch}
        \State $O_{W32A32} \leftarrow$ \Call{forward}{$\mathcal{M}_{W32A32}$, batch}
        \State $l \leftarrow$ \Call{$\mathcal{L}$}{$\mathrm{stop\_grad}(O_{W32A32})$, $O_{WbA8}$}
        \State \Call{optimize}{$\mathcal{O}$, $l$, $\mathcal{M}_{WbA8}$}

        \EndFor
        
        \EndFor \\
        \Return $\mathcal{M}_{WbA8}$
    \end{algorithmic}
\end{algorithm}

\section{Related works} \label{sec:related}

Large diffusion models are the de-facto framework for image generation~\cite{dhariwal2021diffusion,rombach2022high,flow_matching,gat2024discrete}. On the other hand, Large Language Models (LLMs) shows human-like abilities with text~\cite{grattafiori2024llama,achiam2023gpt,team2023gemini}. However quality and diversity comes at a cost: these models have a huge amount of parameters and cannot be hosted on an on-device NPU without some form of quantization.

\paragraph{Tackling computational complexity}
Diffusion models' computational complexity has two major sources: the amount of denoising passes and the conditioning mechanism. The former issue can be addressed by distilling the model to few or a single pass~\cite{salimans2022progressive,noroozi2025you}, while the latter by modeling the latent-noise space~\cite{noroozi2025guidance} to decrease the number of function evaluations. Despite the success of these approaches, a major bottleneck remains: the memory required for inference. Quantization %
aim to preserve the original model's quality when moving to lower precision -- thus saving memory and enabling deployment.

\paragraph{Quantization-Aware-Training}%
QAT methods optimize model's weights at lower precision~\cite{bulat2021bit,jin2020adabits} aiming to recover the original performance. Early approaches \cite{bulat2021bit,jin2020adabits} study QAT on ResNet for classification: starting from low bit precision -- $b=2 \text{ or } 4$ -- the weights quantization is progressively reduced~\cite{jin2020adabits} or selected at random\cite{bulat2021bit}. Although \cite{bulat2021bit,jin2020adabits} closely relate to \short, they (i) focus on classification networks, (ii) ignore the gap with full precision models and the hierarchal nature of different precisions by starting from a low-bit quantization, (iii) aim to get models at different precisions. In parallel, Fracbits~\cite{yang2021fracbits} introduces bit-width optimization by relying on a non-standard quantization formula for fractional bits. Bit-width are regularized to achieve the desired precision, followed by a binary search and fine tuning process to finalize weights. However, this procedure focuses on average bit length across the layers, obtaining lower bits in some layers at the cost of higher bits in other layers, which may not map to readily available hardware. Finally, as \cite{bulat2021bit,jin2020adabits}, Fracbits focuses on classification rather than generative tasks.

More recently, MatryoshkaQAT~\cite{nair2025matryoshka} exploits the nested structure of a number's byte representation to encode LLM's weights at different precision -- $8, 4\text{, and }2$. The joint training at the three precisions result in a multi-precision model. Parallel to this work, Liu \etal~\cite{liu2025paretoq} extend \cite{nagel2022overcoming}, and discover that model quantized to lower than 4 bits develop a different representation from the original models. Finally, \cite{liu2022bit} based on the Teaching Assistant distillation framework \cite{mirzadeh2020improved}, quantize LLM models to \WA{1}{1}. Similar to our work, the authors use a progressive strategy, however limited to \WA{1}{4} (to \WA{1}{2}) to \WA{1}{1}, where intermediate models (\WA{1}{2}) models are used as teachers. Combined with a series of techniques (\eg, gradient clipping, elastic binarization, etc) to stabilize the optimization process the authors achieve a binary quantized model. Since the quantization is binary, the model is not deployable to edge-devices. In this work, we show progressive quantization is enough to quantize a model that can be deployed on edge-devices.

Related to diffusion models, Bitfusion~\cite{sui2024bitsfusion} combines different QAT techniques, such as distillation and fine tuning, to convert SD1.5~\cite{rombach2022high} to 1.99 bits. Following a similar trend, Wang \etal~in~\cite{wang2024quest} selectively fine tune SD1.5 to handle activation distribution. BinaryDM~\cite{zheng2024binarydm} takes model quantization one step further by applying a multi-stage QAT approach to quantization. Notwithstanding these impressive results, none of these techniques showcase low-bit quantization of large scale DiT models such as SD3.5-M~\cite{sd352024} (2.2B) or FLUX.1-schnell~\cite{flux2024} (12B). Ours is the first QAT approach applied to such models.

\paragraph{Post-Training Quantization}%
Among the recent seminal works on LLMs, SmoothQuant~\cite{smooth_quant} proposes a PTQ approach by injecting a smoothing factor in linear layers to reduce the impact of outliers in LLMs. AWQ~\cite{lin2024awq} and MobileQuant~\cite{mobile_quant} extend this approach to lowers the precision to \WA{4}{8} thus enabling an LLM to run on-device.
These works have been extended to DiT models with specific focus on timesteps. PTQ4DiT~\cite{wu2024ptq4dit} builds a calibration dataset by sampling timesteps before quantizing the diffusion model. DiTAS~\cite{ditas} proposes a temporal-aggregated smoothing technique combined with LoRA and a grid-search to reduce quantization errors of small DiT networks with \WA{4}{8} quantization. QuEST~\cite{wang2024quest} through layer specific (PTQ) fine tuning achieves \WA{4}{4} quantization. Q-DiT~\cite{q_dit}, inspired by~\cite{lin2024awq,zhao2024atom,mobile_quant}, combines a fine-grained group quantization with a novel automatic allocation algorithm to account for weights' spatial variance. Most recently, SVDQuant~\cite{li2024svdquant} and FBQuant~\cite{liu2025fbquant} have shown impressive preservation of image quality generation when quantizing FLUX.1-schnell~\cite{flux2024} to \WA{4}{16}. The authors rely on a low-rank approximation of the original weights and a residual branch to absorb outliers.

\section{Experiments} \label{sec:experiments}

\paragraph{Models}
We focus our evaluation on recent text to image (T2I) models since there is an increasing interest in lowering their computational requirements due to their large number of parameters. In particular, we assess the soundness of the proposed approach over 4 different diffusion models: SD3.5-Medium~\cite{sd352024}, Sana~\cite{sana}, \pixart~\cite{chen2024pixart}, and FLUX.1-schnell~\cite{flux2024}. These models space a wide range of parameters, 0.6B--12B, and architectural innovations, linear and non-linear attentions, DiT, MM-DiT, etc.

\begin{table}[t]
    \centering
    \caption{\textbf{Qualitative evaluation:}\label{tab:evaluation_small}
    we evaluate \short{} using a \textit{fractional quantization schedule} on PixArt Evaluation dataset~\cite{chen2024pixart} and  MidJourney HQ Evaluation dataset~\cite{li2024playground} measuring FID, and CLIP-FID wrt the original model, and ImageReward (IR)~\cite{xu2023imagereward}.}
    \resizebox{\columnwidth}{!}{%
    \begin{tabular}{llcccccccccccccccccccc}
        \toprule
        \cols{14}{\pixart} \\
        \toprule
        & & \cols{3}{SD3.5 Medium} & \cols{3}{Sana 600M} & \multicolumn{3}{c}{\pixart} & \multicolumn{3}{c}{Flux-schnell} \\
        \cmidrule(r){3-5} 
        \cmidrule(r){6-8}
        \cmidrule(r){9-11}
        \cmidrule(r){12-14}
        \rows{2}{Method} & \rows{2}{Precision} & \rows{2}{FID $\downarrow$} & CLIP & \rows{2}{IR $\uparrow$} & \rows{2}{FID $\downarrow$} & CLIP & \rows{2}{IR $\uparrow$} & \rows{2}{FID $\downarrow$} & CLIP & \rows{2}{IR $\uparrow$} & \rows{2}{FID $\downarrow$} & CLIP & \rows{2}{IR $\uparrow$} \\
        & & & FID $\downarrow$ & & & FID $\downarrow$ & & & FID $\downarrow$ & & & FID $\downarrow$ & \\
        \toprule
        Dynamic Q. & \WA{4}{8} & 9.36 & 2.08 & 0.56 & 2.22 & 0.24 & 0.57 & 13.35 & 6.19 & 0.35 & 8.17 & 1.13 & -0.73 \\
        DiTAS & \WA{4}{8} & 27.93 & 13.77 & 0.41 & 12.87 & 4.58 & \textbf{0.62} & 7.30 & 3.95 & \textbf{0.84} & - & - & - \\
        SVDQuant & \WA{4}{16} & 14.42 & 3.14 & 0.66 & 2.43 & 0.24 & 0.60 & 6.80 & 2.02 & 0.79 &  \textbf{2.26} & 0.36 & 0.84 \\
        \midrule
        SVDQAT & \WA{4}{8} & 2.57 & 0.28 & 0.80 & \textbf{1.93} & \textbf{0.13} & 0.48 & 5.38 & 1.48 & 0.76 & - & - & - \\ 
        vQAT & \WA{4}{8} & 2.67 & 0.31 & 0.78 & 2.13 & 0.16 & 0.45 & 7.00 & 2.52 & 0.79 & 3.40 & 0.66 & \textbf{0.87} \\ 
        \rowcolor{lightblue} \textbf{\short} & \WA{4}{8} & \textbf{2.54} & \textbf{0.27} & \textbf{0.82} & 2.17 & 0.19 & 0.48 & \textbf{4.48} & \textbf{1.07} & 0.79 & 2.55 & \textbf{0.30} & 0.86 \\
        
        \toprule
        \cols{14}{MJHQ} \\
        \toprule
        & & \cols{3}{SD3.5 Medium} & \cols{3}{Sana 600M} & \multicolumn{3}{c}{\pixart} & \multicolumn{3}{c}{Flux-schnell} \\
        \cmidrule(r){3-5} 
        \cmidrule(r){6-8}
        \cmidrule(r){9-11}
        \cmidrule(r){12-14}
        \rows{2}{Method} & \rows{2}{Precision} & \rows{2}{FID $\downarrow$} & CLIP & \rows{2}{IR $\uparrow$} & \rows{2}{FID $\downarrow$} & CLIP & \rows{2}{IR $\uparrow$} & \rows{2}{FID $\downarrow$} & CLIP & \rows{2}{IR $\uparrow$} & \rows{2}{FID $\downarrow$} & CLIP & \rows{2}{IR $\uparrow$} \\
        & & & FID $\downarrow$ & & & FID $\downarrow$ & & & FID $\downarrow$ & & & FID $\downarrow$ & \\
        \toprule
        Dynamic Q. & \WA{4}{8} & 10.29 & 2.11 & 0.65 & 2.40 & 0.28 & 0.63 & 15.04 & 5.55 & 0.44 & 8.66 & 1.24 & -0.90 \\
        DiTAS & \WA{4}{8} & 32.04 & 14.06 & 0.41 & 12.91 & 5.59 & \textbf{0.68} & 8.63 & 4.07 & \textbf{1.04} & - & - & - \\
        SVDQuant & \WA{4}{16} & 15.10 & 3.06 & 0.78 & 2.48 & 0.25 & 0.62 & 6.95 & 1.71 & 0.99 & \textbf{2.41} & 0.41 & 0.96 \\
        \midrule
        SVDQAT & \WA{4}{8} & 2.85 & \textbf{0.32} & 0.91 & \textbf{2.04} & \textbf{0.16} & 0.53 & 5.83 & 1.44 & 0.96 & - & - & - \\ 
        vQAT & \WA{4}{8} & 3.01 & 0.37 & 0.89 & 2.13 & 0.20 & 0.47 & 7.38 & 2.12 & 0.99 & 3.56 & 0.73 & \textbf{0.99} \\ 
        \rowcolor{lightblue} \textbf{\short} & \WA{4}{8} & \textbf{2.78} & \textbf{0.32} & \textbf{0.96} & 2.34 & 0.24 & 0.50 & \textbf{4.95} & \textbf{1.05}4 & 0.97 & 2.55 & \textbf{0.39}& \textbf{0.99} \\
        \midrule
    \end{tabular}%
    }
    \vspace{-0.5cm}
\end{table}

In all our experiments, we start from a pre-trained \WA{32}{32} model and through \short{} reduce it to \WA{4}{8}. We bootstrap the student at \INT{8} and optimize it to replicate its \FP{32} counter-part. This initialization allows \short{} to start with minimal gap between teacher (\FP{32}) and student (\INT{8}). After $\T$ epochs we lower the precision of the model -- number of bits --, and continue the optimization. This procedure is repeated until the precision reaches $4$ bits. Since we apply a \textit{fake quantization} process, we can emulate arbitrary precisions that have no hardware support, \eg, \INT{4.5}.
Unless stated otherwise all our experiments follow the same progression: $8 \rightarrow 7 \rightarrow 6 \rightarrow 5.5 \rightarrow 5 \rightarrow 4.75 \rightarrow 4.5 \rightarrow 4.25 \rightarrow 4$ targeting linear layers. Further hyper-parameters are detailed in the appendix. In all cases we distil a \WA{4}{8} model through knowledge-distillation loss, using dynamic quantization. Note that the proposed approach is applicable to any quantization precision, \eg, \WA{2}{8}, and static quantization.

\paragraph{Baselines}
We compare the proposed approach with state of the art PTQ techniques: DiTAS~\cite{ditas} (\WA{4}{8}) and SVDQuant~\cite{li2024svdquant}(\WA{4}{16}). In both cases, we use the code publicly available and train (calibrate) the model over the train dataset (see below). To further prove the soundness of \short{}, we implement a vanilla QAT (vQAT) approach, and an SVDQuant-like QAT (SVDQAT). In vQAT, we apply a \WA{4}{8} quantization to all linear layers and optimize them with the same loss as \short{}. Instead in SVDQAT, we inject a LoRA-like layer in all linear layers, as in ~\cite{li2024svdquant}, and optimize both the low-rank and residual branch -- almost doubling the number of parameters. Finally, we report the results for the naive quantization (Dynamic Q.) to desired precision through torchao\footnote{\url{https://github.com/pytorch/ao}}.

\paragraph{Datasets} 
All models in all our experiments are trained -- calibrated -- on YE-POP\footnote{\url{https://huggingface.co/datasets/Ejafa/ye-pop}} dataset. We split it between training ($97.5\%$) and validation ($2.5\%$). Then, quantized models are evaluated on two different datasets: \pixart~Evaluation dataset~\cite{chen2024pixart}\footnote{\url{https://huggingface.co/datasets/PixArt-alpha/PixArt-Eval-30K}}, and MidJourney HQ Evaluation dataset~\cite{li2024playground}. In all cases during training and evaluation, we generate $512\times512$ images.

\paragraph{Metrics}
We quantitatively evaluate the proposed technique over a variety of metrics measuring image quality, and features distributions. In particular, we measure the image quality with Image Reward (IR)~\cite{xu2023imagereward}, and measure the features distribution disparity between the generated samples of the quantized model and the original one with FID~\cite{szegedy2016rethinking} and CLIP-FID~\cite{kynkaanniemi2022role}. This choice allows us to quantify the similarity between the original model and its quantized version: similar images have similar features -- thus lower FID score.

\subsection{Quantitative evaluation} \label{sec:quantitative}

\tabref{tab:evaluation_small}
shows a quantitative comparison of \short{} across the aforementioned five models, two SOTA QAT approaches for direct comparison, and three PTQ techniques for overall completeness across all quantization approaches. All combinations are evaluated across two different test datasets. Due to memory requirements, we are unable to apply some techniques to Flux-schnell~\cite{flux2024} (12B model).

SVDQuant was developed and optimized for Sana, \pixart{} and FLUX.1-schnell, thus achieves lower performance in SD3.5-Medium, as shown in \tabref{tab:evaluation_small}, yielding particularly worse FID and CLIP FID metrics for both test datasets.
\tabref{tab:evaluation_small} reveals mixed results for Dynamic Quantization and DiTaS on both test datasets. Specifically, DiTaS outperforms Dynamic Quantization in \pixart{} but reveals itself overall worse for SD3.5-Medium and Sana. These models have different architectures, namely DiT, MM-DiT and linear attention, which indicates that DiTaS can be particularly sensitive to the model family.

As for other QAT approaches, our developed SVDQuant-like QAT (SVDQAT) consistently outperforms vanilla QAT (vQAT) across the test datasets. Arguably, the increased number of parameters -- LoRA and residual branch -- better cope with the lower precision. Our motivation to establish this strong QAT baseline was the success of SVDQuant as a PTQ approach alongside its need of a higher precision \WA{4}{16} that hinders latency in on-device accelerators.
Finally, \tabref{tab:evaluation_small} shows \short{} outperforms even the strongest QAT baseline we developed namely SVDQAT, with overall higher gains for SD3.5-Medium and \pixart{} for both test datasets.

\begin{figure}[ht!]
    \centering
    \setlength{\tabcolsep}{0.0pt}
    \renewcommand{\arraystretch}{0}%
    \begin{tabular}{ccccc}
        \multicolumn{4}{c}{\textbf{\LARGE SD3.5-M}} \\[2pt]
        \includegraphics[width=0.25\textwidth,height=0.25\textwidth]{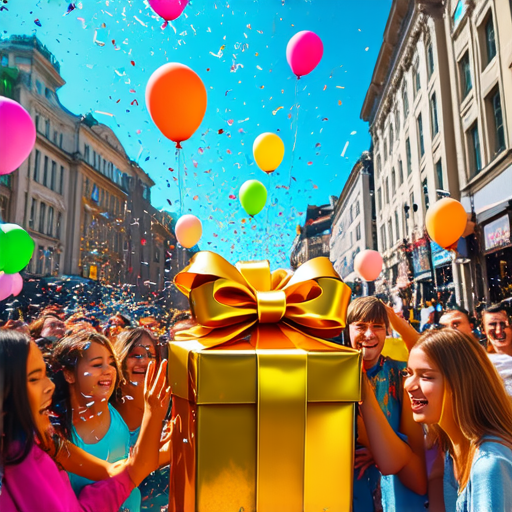} &
        \includegraphics[width=0.25\textwidth,height=0.25\textwidth]{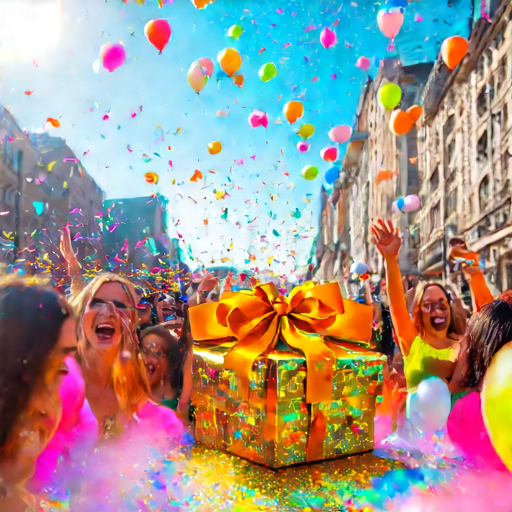} &
        \includegraphics[width=0.25\textwidth,height=0.25\textwidth]{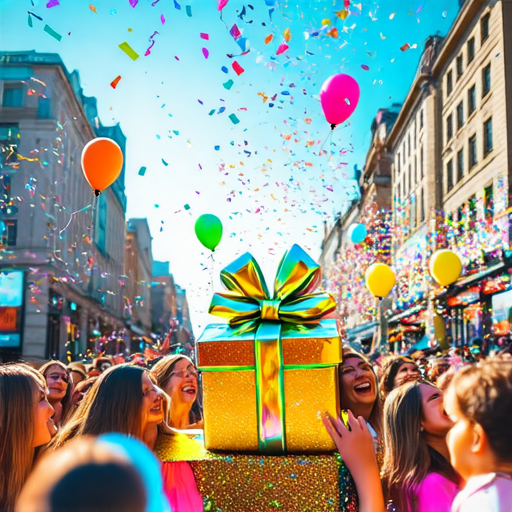} &
        \includegraphics[width=0.25\textwidth,height=0.25\textwidth]{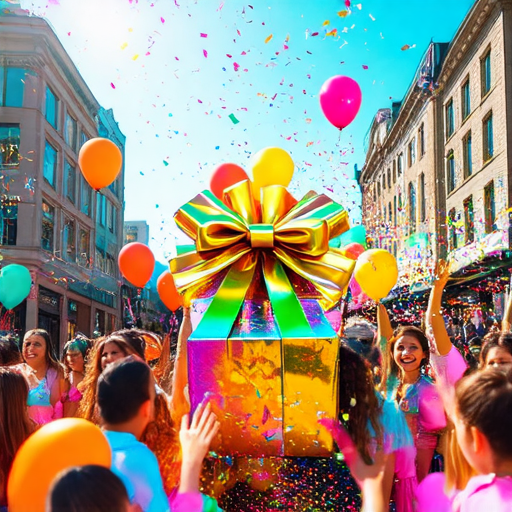} \\
        [2pt]
        \multicolumn{4}{c}{\textbf{\LARGE Sana 600M}} \\[2pt]
        \includegraphics[width=0.25\textwidth,height=0.25\textwidth]{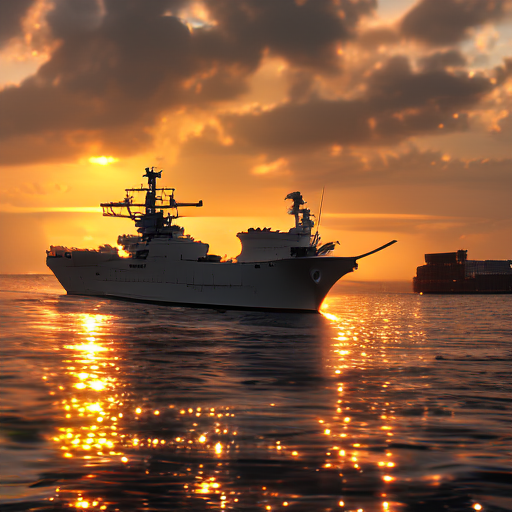} &
        \includegraphics[width=0.25\textwidth,height=0.25\textwidth]{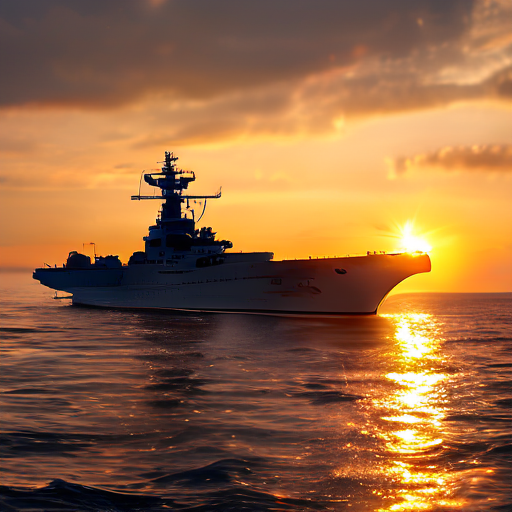} &
        \includegraphics[width=0.25\textwidth,height=0.25\textwidth]{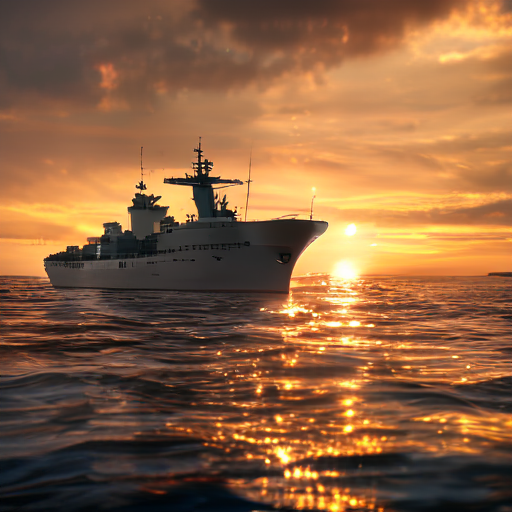} &
        \includegraphics[width=0.25\textwidth,height=0.25\textwidth]{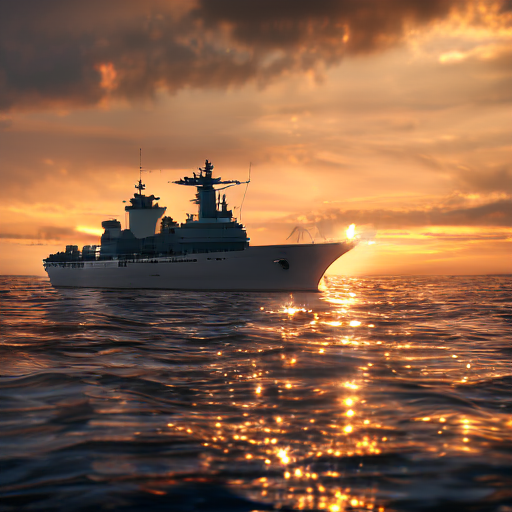} \\
        [2pt]
        \multicolumn{4}{c}{\textbf{\LARGE \pixart}} \\[2pt]
        \includegraphics[width=0.25\textwidth,height=0.25\textwidth]{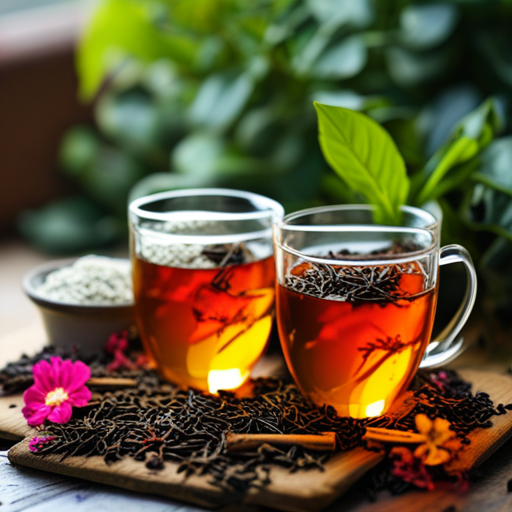} &
        \includegraphics[width=0.25\textwidth,height=0.25\textwidth]{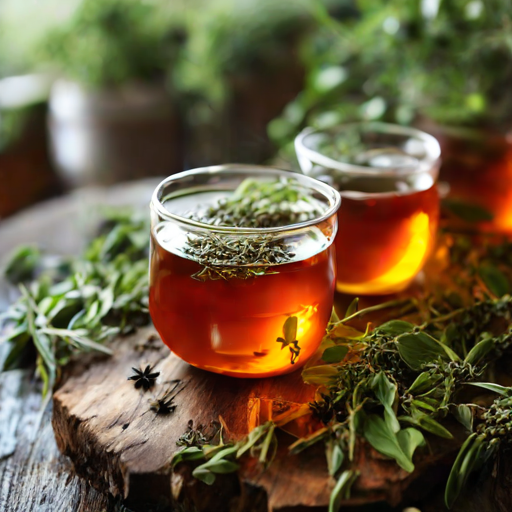} &
        \includegraphics[width=0.25\textwidth,height=0.25\textwidth]{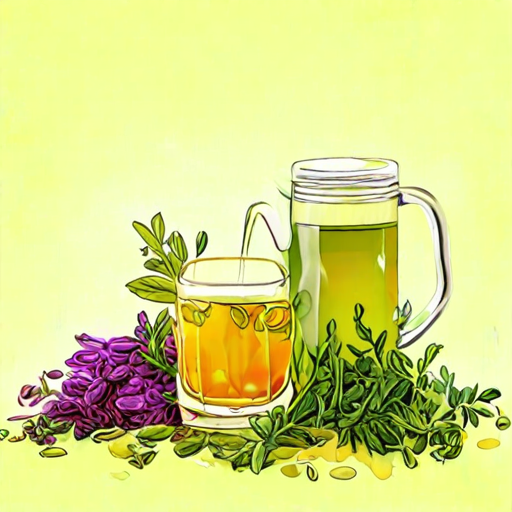} &
        \includegraphics[width=0.25\textwidth,height=0.25\textwidth]{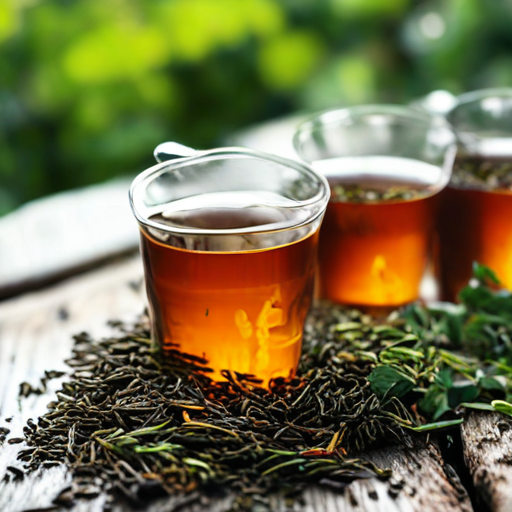} \\
        [2pt]
        \multicolumn{4}{c}{\textbf{\LARGE Flux-Schnell}} \\[2pt]
        \includegraphics[width=0.25\textwidth,height=0.25\textwidth]{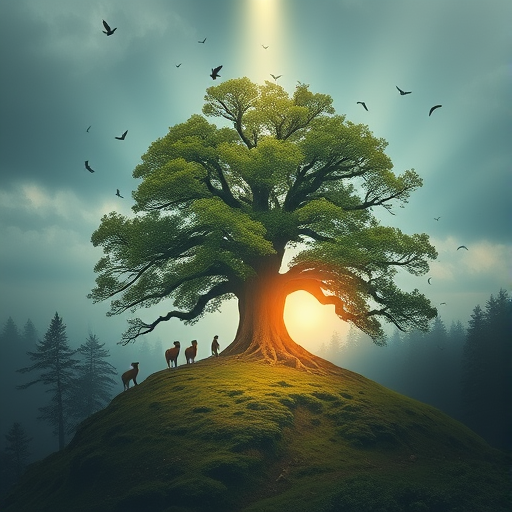} &
        \includegraphics[width=0.25\textwidth,height=0.25\textwidth]{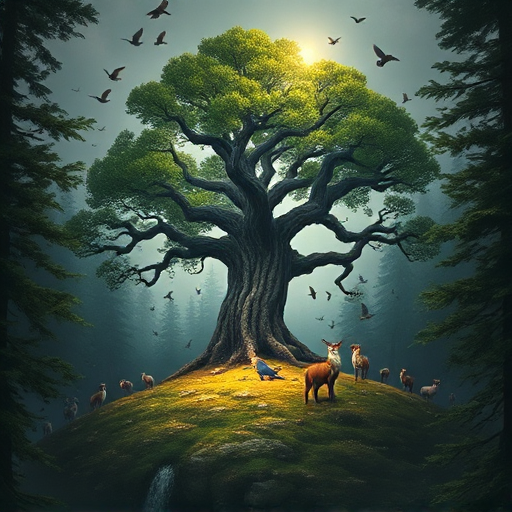} &
        \includegraphics[width=0.25\textwidth,height=0.25\textwidth]{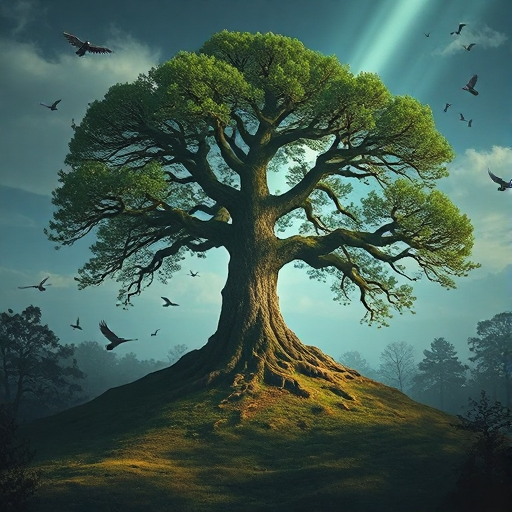} &
        \includegraphics[width=0.25\textwidth,height=0.25\textwidth]{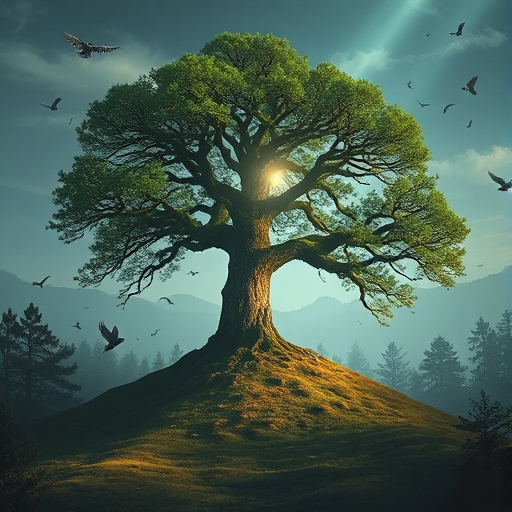} \\
        \includegraphics[width=0.25\textwidth,height=0.25\textwidth]{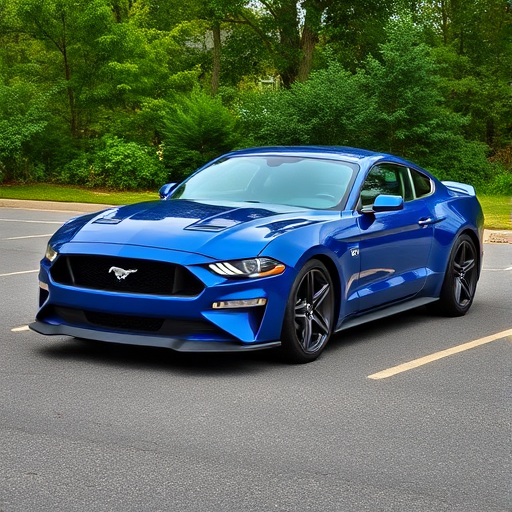} &
        \includegraphics[width=0.25\textwidth,height=0.25\textwidth]{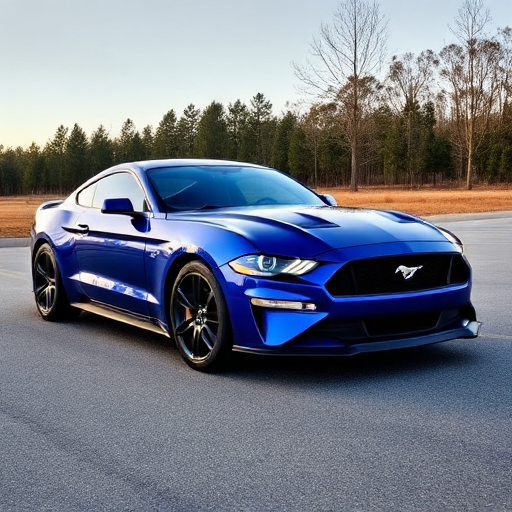} &
        \includegraphics[width=0.25\textwidth,height=0.25\textwidth]{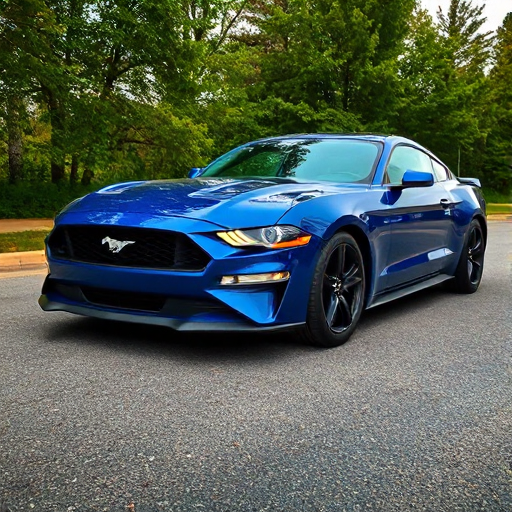} &
        \includegraphics[width=0.25\textwidth,height=0.25\textwidth]{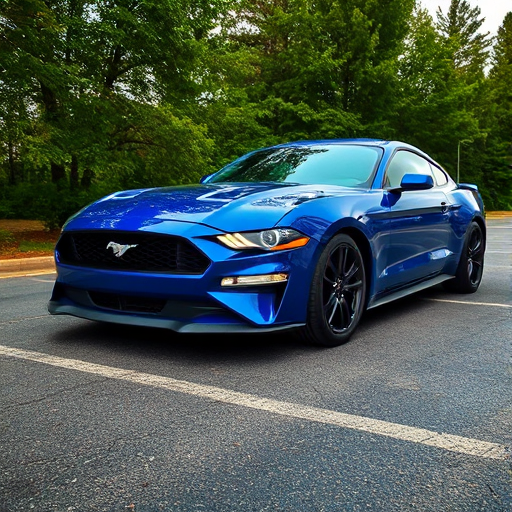} \\
        [2pt]
        \textbf{(a)} Original model &
        \textbf{(b)} SVDQuant &
        \textbf{(c)} vQAT &
        \textbf{(d)} \short{} (Ours) \\
         
    \end{tabular}
    \caption{\textbf{Qualitative comparison:}\label{fig:qualitative}
    \short (d) generates images similar to the original model (a). Prompts are from MJHQ dataset~\cite{li2024playground}.}
\end{figure}

\subsection{Qualitative evaluation} \label{sec:qualitative}

To complement the quantitative evaluation in the preceding section, \figref{fig:qualitative} depicts for each of the four models considered: the original model, one PTQ representative (SVDQuant), one QAT alternative (vQAT) and the proposed QAT approach (\short). We have selected SVDQuant among DiTaS and Dynamic Quantization given its popularity. We focus on vQAT rather than our other developed QAT baseline (SVDQAT) given its overall popularity. Note that, the images generated in each row in \figref{fig:qualitative} share the same seed and prompt.

As expected from a PTQ approach, SVDQuant under-performs when generating certain high frequency image details. This is especially visible when multiple faces are present in a generated image as shown in the first row. QAT improves high frequency image details, but generates significantly different images than the original model for the same prompt and seed. Finally, \short{} preserves both high frequency details and generates images as close to the original model across all baselines.

\begin{figure}[t]
  \begin{minipage}[t]{0.49\linewidth}
    \centering
    \includegraphics[width=0.99\linewidth]{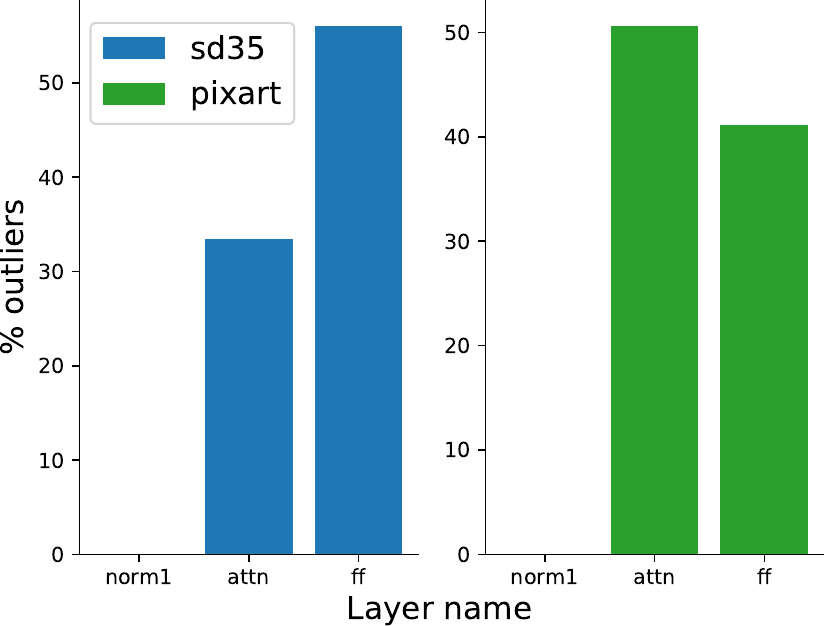}
    \captionof{figure}{\textbf{Outliers:}\label{fig:outliers} outliers distribution for activations varies across models. SD3.5-M (left) experience most of its outliers right after Feed Forward layers, while for \pixart, most outliers are in Attention layers.}
  \end{minipage}\hfill
  \begin{minipage}[b]{.49\linewidth}
    \centering
    \captionof{table}{\textbf{Outlier analysis:} \label{tab:layer_ablation} we optimize specific layers types while the rest of the model is frozen and quantized (\WA{4}{8}). FID and CLIP-FID are computed on \pixart~\cite{chen2024pixart} evaluation dataset.}

    \begin{tabular}{llcc}
        \toprule
        Model & Layer & FID $\downarrow$ & CLIP FID $\downarrow$ \\
        \toprule
        \rows{4}{SD3.5-M} & \cfirst FF  & \cfirst \textbf{2.23} & \cfirst 0.23 \\
                          & Attn & 2.32 & 0.24 \\
                          & TF   & 2.49 & 0.28 \\
                          & All  & 2.54 & \textbf{0.22} \\
        \midrule
        \rows{4}{Sana 600M} & FF   & 2.18 & 0.17 \\
                    & \cfirst Attn & \cfirst \textbf{2.10} & \cfirst \textbf{0.16} \\ 
                            & TF   & 2.13 & \textbf{0.16} \\
                            & All  & 2.17 & 0.19 \\
        \midrule
        \rows{4}{\pixart} & FF   & 5.34 & 1.55 \\
                          & Attn & 6.48 & 2.23 \\ 
                    & \cfirst TF & \cfirst \textbf{4.40} & \cfirst \textbf{1.13} \\
                          & All  & 4.48 & 1.07 \\
        \midrule
    \end{tabular}
    
  \end{minipage}
\end{figure}

\subsection{Outlier analysis} \label{sec:outliers}

Activation outliers disrupt the quantization process by introducing artifacts or biases. By analyzing these outliers across different models, we discover that different models produce outliers in different layers. For example, in SD3.5-M outliers emerge after Feed-Forward (FF) layers, while in \pixart{} outliers arise from Attention (Attn) layers, see \figref{fig:outliers}. Through selectively training specific layers, we can \textit{reduce \short's computational demand} while obtaining a deployable model. In this vein, we analyze the impact of selective training, \ie, we optimize only certain layers while the rest of the network is \textit{frozen and quantized} (\WA{4}{8}). In particular, we focus on attention layers (Attn), feed forward layers (FF), and transformer blocks (TF), and compare it with training the entire network (Full).

Quantitative results in \tabref{tab:layer_ablation} show that there is no clear winner -- a layer type for all architecture. Different models take advantage from optimizing different layers. Nevertheless we recommend starting from quantizing Transformer Blocks (TF) as it reduces memory requirements, lowers computational demands, and addresses all outliers.

\begin{figure}[t]
    \centering
    \setlength{\tabcolsep}{0.0pt}
    \begin{tabular}{ccc}
        \includegraphics[width=0.33\linewidth]{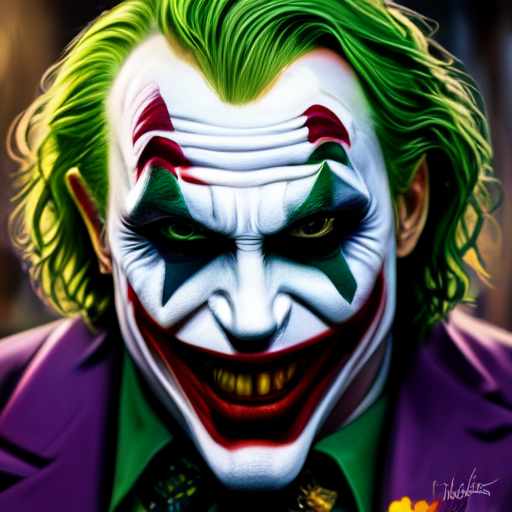} &
        \includegraphics[width=0.33\linewidth]{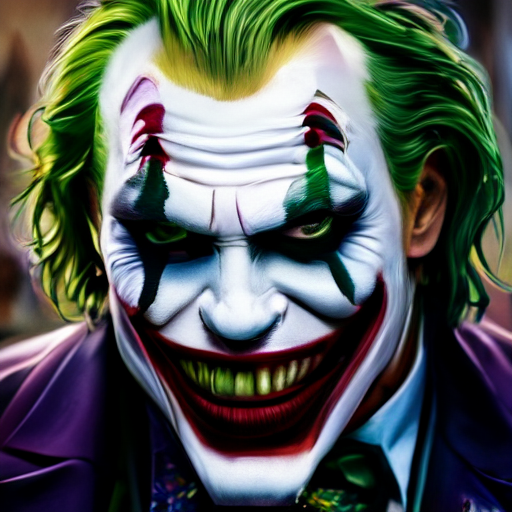} &
        \includegraphics[width=0.33\linewidth]{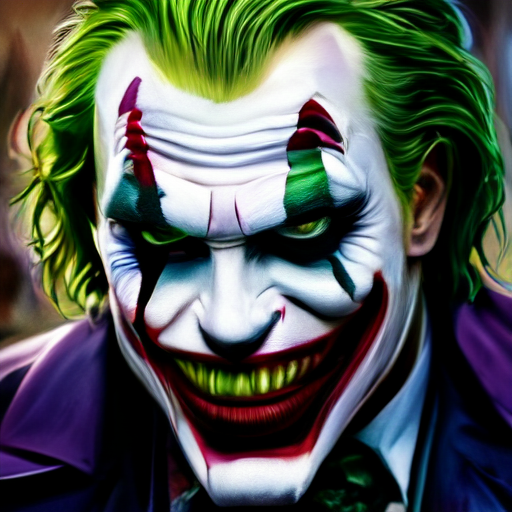} \\
        \includegraphics[width=0.33\linewidth]{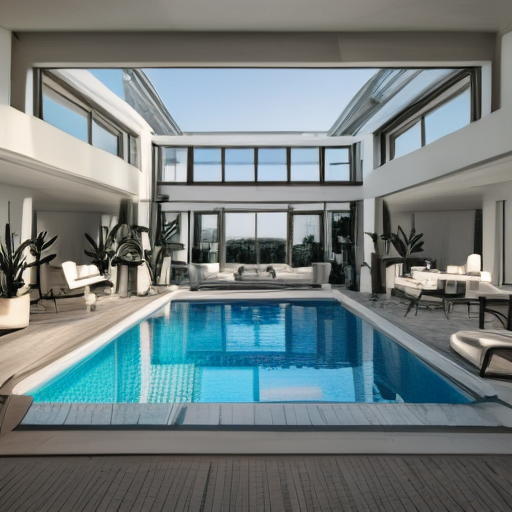} &
        \includegraphics[width=0.33\linewidth]{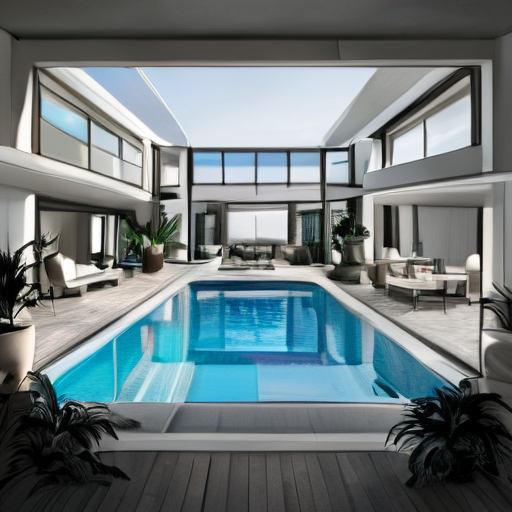} &
        \includegraphics[width=0.33\linewidth]{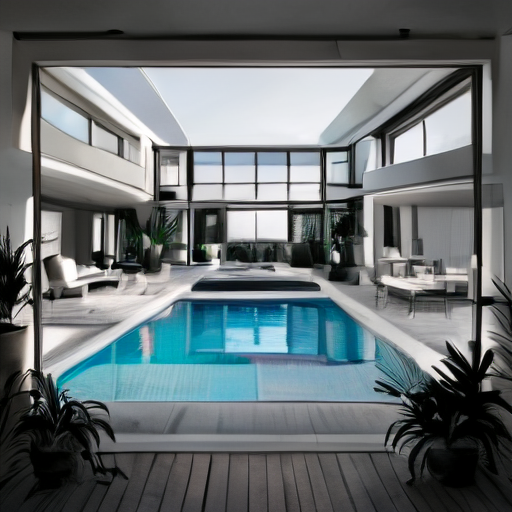} \\
        \textbf{(a)} Original Sana 600M & 
        \textbf{(b)} \short{} on GPU & 
        \textbf{(c)} \short{} on smartphone \\
    \end{tabular}
    \caption{\textbf{On device generation:}\label{fig:on_device} we generate images on a mobile phone (c) and compare the results with the samples generates on GPU by the original model (a) and the quantized model (b).}
\end{figure}

\subsection{On device model deployment}\label{sec:on_device}

To demonstrate the feasibility of deploying models quantized with \short{} on edge devices, we quantized Sana 600M~\cite{sana} to \WA{4}{8} and deployed it on the Samsung S25U, which runs on the Qualcomm SM8750-AB Snapdragon 8 Elite Hexagon Tensor Processor (HTP). Compared to CPUs and GPUs, integer accelerators support a limited range of precisions and exclusively support static quantization for both weights and activations. Please, refer to \secref{sec:preliminaries}, for a discussion on static and dynamic quantization.

Conversely to the baselines used in \secref{sec:quantitative}, \short{} support both dynamic and static quantization paradigms. To apply \short{} to Sana 600M~\cite{sana} with static weight and activation quantization, we pre-compute scale and offset through statistical analysis of the features: we select 100 random samples and pass them through the DiT. Feature values for every layer are recorded and used to compute standard deviation and mean. Following literature~\cite{wang2025optimizing}, we use a 3 std range for inlier. Finally, we use these scale and offset during the QAT, while the overall training procedure does not change from the one discussed in \secref{sec:quantitative}.

All linear layers of the quantized model run with precision \WA{4}{8} except for the last layer which runs in \WA{4}{16}. This is a good compromise to preserve quality, while not impacting latency. Overall, the model has a latency of $66ms$ per forward step, while the same model running in \WA{4}{16} (bit-width supported by SVDQuant) has a latency of $95ms$ -- a $30.5\%$ latency improvement. Finally, to assess the on-device quality we generate samples, and compare them with the those from the original model in \figref{fig:on_device}. The quantized model produces high quality pictures that resemble the original model and its GPU version.

\section{Limitations and future work} \label{sec:limitations}

The proposed approach is a step forward compared to SOTA Quantization-Aware-Training (QAT).
Like most -- if not all -- QAT techniques, \short{} yields higher quality but is more computational expensive than Post-Training Quantization (PTQ). Compared with multi-precision LLM's SOTA QAT approaches, such as MatryoshkaQAT, \short's quantized model is tailored to a single bit precision, and we leave multi-precision support for future work. Furthermore, the intermediate precision levels are hand-picked, in the future, we plan to design an algorithm to select the most impactful precisions.

The proposed training scheme may benefit from regularizers such as weight decay and data augmentation. For example, preliminary regularization tests on Sana 0.6B~\cite{sana} show that weight decay boosts the performance by $\sim 10\%$. A proper investigation of regularization and its impact in training is left to future work.
Finally, \short's networks are optimized using knowledge distillation only, but different losses such as feature and task loss may further boost image generation quality.

\section{Conclusions}

\short{} is a novel Quantization-Aware-Training technique that exploits fractional bits while progressively reducing the parameter-precision during the quantization process. Thanks to this curriculum learning strategy, we address the outliers, as they arise, at different precisions, achieving a more stable and faster training. The proposed method is evaluated over a variety of state-of-the-art DiT and MM-DiT models. We show that, both qualitatively and quantitatively, the quantized models achieve superior performance compared to the state-of-the-art QAT approaches. Such improved quality, if deployed on-device, may boost mobile users productivity, preserve their privacy, as well as generate personalized contents for users.

\bibliography{references}

\begin{thebibliography}{10}

\bibitem{sd352024}
S.~AI, ``Stable diffusion 3.5.'' \url{https://stability.ai/news/introducing-stable-diffusion-3-5}, 2024.

\bibitem{sana_15}
E.~Xie, J.~Chen, Y.~Zhao, J.~Yu, L.~Zhu, C.~Wu, Y.~Lin, Z.~Zhang, M.~Li, J.~Chen, H.~Cai, B.~Liu, D.~Zhou, and S.~Han, ``{SANA 1.5: Efficient Scaling of Training-Time and Inference-Time Compute in Linear Diffusion Transformer},'' {\em arXiv:2501.18427}, 2025.

\bibitem{flux2024}
B.~F. Labs, ``Flux.'' \url{https://github.com/black-forest-labs/flux}, 2024.

\bibitem{grattafiori2024llama}
A.~Grattafiori, A.~Dubey, A.~Jauhri, A.~Pandey, A.~Kadian, A.~Al-Dahle, A.~Letman, A.~Mathur, A.~Schelten, A.~Vaughan, {\em et~al.}, ``The llama 3 herd of models,'' {\em arXiv preprint arXiv:2407.21783}, 2024.

\bibitem{team2025gemma}
G.~Team, A.~Kamath, J.~Ferret, S.~Pathak, N.~Vieillard, R.~Merhej, S.~Perrin, T.~Matejovicova, A.~Ram{\'e}, M.~Rivi{\`e}re, {\em et~al.}, ``Gemma 3 technical report,'' {\em arXiv preprint arXiv:2503.19786}, 2025.

\bibitem{qualcomm_snapdragon}
Q.~Snapdragon\textregistered, ``{Snapdragon 8 Elite Mobile Platform}.'' \url{https://docs.qualcomm.com/bundle/publicresource/87-83196-1_REV_D_Snapdragon_8_Elite_Mobile_Platform_Product_Brief.pdf}.

\bibitem{intel_xeon}
X.~Intel\textregistered, ``{Processors with Performance-Cores (P-Cores)}.'' \url{https://www.intel.com/content/www/us/en/products/details/processors/xeon/xeon6-p-cores.html}.

\bibitem{amd_ryzen}
A.~Ryzen\texttrademark, ``{AI 300 Series Processors}.'' \url{https://www.amd.com/content/dam/amd/en/documents/partner-hub/ryzen/amd-ryzen-ai-300-series-vs-qualcomm-snapdragon-x-elite-deck.pdf}.

\bibitem{nvidia_hgx}
N.~HGX\texttrademark, ``{NVIDIA HGX Platform}.'' \url{https://www.nvidia.com/en-gb/data-center/hgx}.

\bibitem{sana}
E.~Xie, J.~Chen, J.~Chen, H.~Cai, H.~Tang, Y.~Lin, Z.~Zhang, M.~Li, L.~Zhu, Y.~Lu, and S.~Han, ``{SANA: Efficient High-Resolution Image Synthesis with Linear Diffusion Transformers},'' {\em arXiv:2410.10629}, 2024.

\bibitem{sana_sprint}
J.~Chen, S.~Xue, Y.~Zhao, J.~Yu, S.~Paul, J.~Chen, H.~Cai, E.~Xie, and S.~Han, ``{SANA-Sprint: One-Step Diffusion with Continuous-Time Consistency Distillation},'' {\em arXiv:2503.09641}, 2025.

\bibitem{li2024svdquant}
M.~Li, Y.~Lin, Z.~Zhang, T.~Cai, X.~Li, J.~Guo, E.~Xie, C.~Meng, J.-Y. Zhu, and S.~Han, ``{SVDQuant: Absorbing Outliers by Low-Rank Components for 4-Bit Diffusion Models},'' {\em arXiv:2411.05007}, 2024.

\bibitem{zhang2025selectq}
Z.~Zhang, Y.~Gao, J.~Fan, Z.~Zhao, Y.~Yang, and S.~Yan, ``Selectq: Calibration data selection for post-training quantization,'' {\em Machine Intelligence Research}, pp.~1--12, 2025.

\bibitem{nair2025matryoshka}
P.~Nair, P.~Datta, J.~Dean, P.~Jain, and A.~Kusupati, ``{Matryoshka Quantization},'' {\em arXiv:2502.06786}, 2025.

\bibitem{bulat2021bit}
A.~Bulat and G.~Tzimiropoulos, ``{Bit-Mixer: Mixed-precision networks with runtime bit-width selection},'' in {\em ICCV}, pp.~5188--5197, 2021.

\bibitem{sui2024bitsfusion}
Y.~Sui, Y.~Li, A.~Kag, Y.~Idelbayev, J.~Cao, J.~Hu, D.~Sagar, B.~Yuan, S.~Tulyakov, and J.~Ren, ``{Bitsfusion: 1.99 bits weight quantization of diffusion model},'' {\em arXiv:2406.04333}, 2024.

\bibitem{bengio2009curriculum}
Y.~Bengio, J.~Louradour, R.~Collobert, and J.~Weston, ``Curriculum learning,'' in {\em Proceedings of the 26th annual international conference on machine learning}, pp.~41--48, 2009.

\bibitem{dhariwal2021diffusion}
P.~Dhariwal and A.~Nichol, ``Diffusion models beat gans on image synthesis,'' {\em Advances in neural information processing systems}, vol.~34, pp.~8780--8794, 2021.

\bibitem{rombach2022high}
R.~Rombach, A.~Blattmann, D.~Lorenz, P.~Esser, and B.~Ommer, ``{High-Resolution Image Synthesis with Latent Diffusion Models},'' in {\em CVPR}, pp.~10684--10695, 2022.

\bibitem{flow_matching}
Y.~Lipman, R.~T.~Q. Chen, H.~Ben-Hamu, M.~Nickel, and M.~Le, ``{Flow Matching for Generative Modeling},'' {\em arXiv:2210.02747}, 2023.

\bibitem{gat2024discrete}
I.~Gat, T.~Remez, N.~Shaul, F.~Kreuk, R.~T. Chen, G.~Synnaeve, Y.~Adi, and Y.~Lipman, ``Discrete flow matching,'' {\em Advances in Neural Information Processing Systems}, vol.~37, pp.~133345--133385, 2024.

\bibitem{achiam2023gpt}
J.~Achiam, S.~Adler, S.~Agarwal, L.~Ahmad, I.~Akkaya, F.~L. Aleman, D.~Almeida, J.~Altenschmidt, S.~Altman, S.~Anadkat, {\em et~al.}, ``Gpt-4 technical report,'' {\em arXiv preprint arXiv:2303.08774}, 2023.

\bibitem{team2023gemini}
G.~Team, R.~Anil, S.~Borgeaud, J.-B. Alayrac, J.~Yu, R.~Soricut, J.~Schalkwyk, A.~M. Dai, A.~Hauth, K.~Millican, {\em et~al.}, ``Gemini: a family of highly capable multimodal models,'' {\em arXiv preprint arXiv:2312.11805}, 2023.

\bibitem{salimans2022progressive}
T.~Salimans and J.~Ho, ``Progressive distillation for fast sampling of diffusion models,'' {\em arXiv preprint arXiv:2202.00512}, 2022.

\bibitem{noroozi2025you}
M.~Noroozi, I.~Hadji, B.~Martinez, A.~Bulat, and G.~Tzimiropoulos, ``{You Only Need One Step: Fast Super-Resolution with Stable Diffusion via Scale Distillation},'' in {\em ECCV}, pp.~145--161, Springer, 2025.

\bibitem{noroozi2025guidance}
M.~Noroozi, A.~G. Ramos, L.~Morreale, R.~Chavhan, M.~Chadwick, A.~Mehrotra, and S.~Bhattacharya, ``Guidance free image editing via explicit conditioning,'' {\em arXiv preprint arXiv:2503.17593}, 2025.

\bibitem{jin2020adabits}
Q.~Jin, L.~Yang, and Z.~Liao, ``{AdaBits: Neural Network Quantization with Adaptive Bit-Widths},'' in {\em CVPR}, pp.~2146--2156, 2020.

\bibitem{yang2021fracbits}
L.~Yang and Q.~Jin, ``Fracbits: Mixed precision quantization via fractional bit-widths,'' in {\em Proceedings of the AAAI Conference on Artificial Intelligence}, vol.~35, pp.~10612--10620, 2021.

\bibitem{liu2025paretoq}
Z.~Liu, C.~Zhao, H.~Huang, S.~Chen, J.~Zhang, J.~Zhao, S.~Roy, L.~Jin, Y.~Xiong, Y.~Shi, {\em et~al.}, ``{ParetoQ: Scaling Laws in Extremely Low-bit LLM Quantization},'' {\em arXiv:2502.02631}, 2025.

\bibitem{nagel2022overcoming}
M.~Nagel, M.~Fournarakis, Y.~Bondarenko, and T.~Blankevoort, ``{Overcoming Oscillations in Quantization-Aware Training},'' in {\em ICML}, pp.~16318--16330, PMLR, 2022.

\bibitem{liu2022bit}
Z.~Liu, B.~Oguz, A.~Pappu, L.~Xiao, S.~Yih, M.~Li, R.~Krishnamoorthi, and Y.~Mehdad, ``Bit: Robustly binarized multi-distilled transformer,'' {\em Advances in neural information processing systems}, vol.~35, pp.~14303--14316, 2022.

\bibitem{mirzadeh2020improved}
S.~I. Mirzadeh, M.~Farajtabar, A.~Li, N.~Levine, A.~Matsukawa, and H.~Ghasemzadeh, ``Improved knowledge distillation via teacher assistant,'' in {\em Proceedings of the AAAI conference on artificial intelligence}, vol.~34, pp.~5191--5198, 2020.

\bibitem{wang2024quest}
H.~Wang, Y.~Shang, Z.~Yuan, J.~Wu, J.~Yan, and Y.~Yan, ``{QuEST: Low-bit Diffusion Model Quantization via Efficient Selective Finetuning},'' {\em arXiv:2402.03666}, 2024.

\bibitem{zheng2024binarydm}
X.~Zheng, X.~Liu, H.~Qin, X.~Ma, M.~Zhang, H.~Hao, J.~Wang, Z.~Zhao, J.~Guo, and M.~Magno, ``{BinaryDM: Accurate Weight Binarization for Efficient Diffusion Models},'' {\em arXiv:2404.05662}, 2024.

\bibitem{smooth_quant}
G.~Xiao, J.~Lin, M.~Seznec, H.~Wu, J.~Demouth, and S.~Han, ``{SmoothQuant: Accurate and Efficient Post-Training Quantization for Large Language Models},'' {\em arXiv:2211.10438}, 2024.

\bibitem{lin2024awq}
J.~Lin, J.~Tang, H.~Tang, S.~Yang, W.-M. Chen, W.-C. Wang, G.~Xiao, X.~Dang, C.~Gan, and S.~Han, ``{AWQ: Activation-aware Weight Quantization for On-device LLM Compression and Acceleration},'' {\em Proceedings of Machine Learning and Systems}, vol.~6, pp.~87--100, 2024.

\bibitem{mobile_quant}
F.~Tan, R.~Lee, {\L}.~Dudziak, S.~X. Hu, S.~Bhattacharya, T.~Hospedales, G.~Tzimiropoulos, and B.~Martinez, ``{MobileQuant: Mobile-friendly Quantization for On-device Language Models},'' {\em arXiv:2408.13933}, 2024.

\bibitem{wu2024ptq4dit}
J.~Wu, H.~Wang, Y.~Shang, M.~Shah, and Y.~Yan, ``{PTQ4DiT: Post-training Quantization for Diffusion Transformers},'' {\em arXiv:2405.16005}, 2024.

\bibitem{ditas}
Z.~Dong and S.~Q. Zhang, ``{DiTAS: Quantizing Diffusion Transformers via Enhanced Activation Smoothing},'' {\em arXiv:2409.07756}, 2024.

\bibitem{q_dit}
L.~Chen, Y.~Meng, C.~Tang, X.~Ma, J.~Jiang, X.~Wang, Z.~Wang, and W.~Zhu, ``{Q-DiT: Accurate Post-Training Quantization for Diffusion Transformers},'' {\em arXiv:2406.17343}, 2024.

\bibitem{zhao2024atom}
Y.~Zhao, C.-Y. Lin, K.~Zhu, Z.~Ye, L.~Chen, S.~Zheng, L.~Ceze, A.~Krishnamurthy, T.~Chen, and B.~Kasikci, ``Atom: Low-bit quantization for efficient and accurate llm serving,'' {\em Proceedings of Machine Learning and Systems}, vol.~6, pp.~196--209, 2024.

\bibitem{liu2025fbquant}
Y.~Liu, H.~Fang, L.~He, R.~Zhang, Y.~Bai, Y.~Du, and L.~Du, ``{FBQuant: FeedBack Quantization for Large Language Models},'' {\em arXiv:2501.16385}, 2025.

\bibitem{chen2024pixart}
J.~Chen, C.~Ge, E.~Xie, Y.~Wu, L.~Yao, X.~Ren, Z.~Wang, P.~Luo, H.~Lu, and Z.~Li, ``Pixart-$\sigma$: Weak-to-strong training of diffusion transformer for 4k text-to-image generation,'' in {\em European Conference on Computer Vision}, pp.~74--91, Springer, 2024.

\bibitem{li2024playground}
D.~Li, A.~Kamko, E.~Akhgari, A.~Sabet, L.~Xu, and S.~Doshi, ``Playground v2. 5: Three insights towards enhancing aesthetic quality in text-to-image generation,'' {\em arXiv preprint arXiv:2402.17245}, 2024.

\bibitem{xu2023imagereward}
J.~Xu, X.~Liu, Y.~Wu, Y.~Tong, Q.~Li, M.~Ding, J.~Tang, and Y.~Dong, ``Imagereward: Learning and evaluating human preferences for text-to-image generation,'' {\em Advances in Neural Information Processing Systems}, vol.~36, pp.~15903--15935, 2023.

\bibitem{szegedy2016rethinking}
C.~Szegedy, V.~Vanhoucke, S.~Ioffe, J.~Shlens, and Z.~Wojna, ``Rethinking the inception architecture for computer vision,'' in {\em Proceedings of the IEEE conference on computer vision and pattern recognition}, pp.~2818--2826, 2016.

\bibitem{kynkaanniemi2022role}
T.~Kynk{\"a}{\"a}nniemi, T.~Karras, M.~Aittala, T.~Aila, and J.~Lehtinen, ``The role of imagenet classes in fr$\backslash$'echet inception distance,'' {\em arXiv preprint arXiv:2203.06026}, 2022.

\bibitem{wang2025optimizing}
R.~Wang, Y.~Gong, X.~Liu, G.~Zhao, Z.~Yang, B.~Guo, Z.~Zha, and P.~Cheng, ``Optimizing large language model training using fp4 quantization,'' {\em arXiv preprint arXiv:2501.17116}, 2025.

\bibitem{lin2024evaluating}
Z.~Lin, D.~Pathak, B.~Li, J.~Li, X.~Xia, G.~Neubig, P.~Zhang, and D.~Ramanan, ``Evaluating text-to-visual generation with image-to-text generation,'' {\em arXiv preprint arXiv:2404.01291}, 2024.

\bibitem{wang2023exploring}
J.~Wang, K.~C. Chan, and C.~C. Loy, ``Exploring clip for assessing the look and feel of images,'' in {\em Proceedings of the AAAI conference on artificial intelligence}, vol.~37, pp.~2555--2563, 2023.

\bibitem{gemma_2024}
G.~Team, ``Gemma,'' 2024.

\bibitem{raffel2020exploring}
C.~Raffel, N.~Shazeer, A.~Roberts, K.~Lee, S.~Narang, M.~Matena, Y.~Zhou, W.~Li, and P.~J. Liu, ``Exploring the limits of transfer learning with a unified text-to-text transformer,'' {\em Journal of machine learning research}, vol.~21, no.~140, pp.~1--67, 2020.

\bibitem{clark2019boolq}
C.~Clark, K.~Lee, M.-W. Chang, T.~Kwiatkowski, M.~Collins, and K.~Toutanova, ``Boolq: Exploring the surprising difficulty of natural yes/no questions,'' {\em arXiv preprint arXiv:1905.10044}, 2019.

\bibitem{talmor2018commonsenseqa}
A.~Talmor, J.~Herzig, N.~Lourie, and J.~Berant, ``Commonsenseqa: A question answering challenge targeting commonsense knowledge,'' {\em arXiv preprint arXiv:1811.00937}, 2018.

\end{thebibliography}
\bibliographystyle{ieeetr}

\newpage
\appendix

\section{Experimental evaluation}

\subsection{Baselines}
For state of the art baselines we rely on code released by authors\footnote{SVDQuant \url{https://github.com/mit-han-lab/deepcompressor}}\footnote{DiTAS \url{https://github.com/DZY122/DiTAS}} and use the default parameters. For all approaches we use pre-trained models with default resolution $512\times512$. Where needed we change the baselines configurations to use the same model.

\subsection{Hyper-parameters for QAT} \label{sec:params}
We detail the various hyper parameters for all QAT experiments in \tabref{tab:hyperparams}. In all cases we rely on FuseAdam as optimizer and optimize for 25 epochs. All experiments run on AMD MI300X and are implemented using PyTorch\footnote{\url{https://pytorch.org/}}, Lightning\footnote{\url{https://lightning.ai/docs/pytorch/stable/}}, torchao\footnote{\url{https://github.com/pytorch/ao}}, with seed $1234$.

\begin{table}[h]
    \centering
    \caption{\textbf{Hyper-parameters:} \label{tab:hyperparams}
    Detailed hyper-parameters required to replicate all experiments.}
    \resizebox{\columnwidth}{!}{%
    \begin{tabular}{ccccccccccc}
    \toprule
        & & \cols{3}{SD3.5-M} & \cols{3}{Sana 600M} & \multicolumn{3}{c}{\pixart} \\
        \cmidrule(r){3-5} 
        \cmidrule(r){6-8}
        \cmidrule(r){9-11}
        & & \rows{2}{lr} & batch & low & \rows{2}{lr} & batch & low & \rows{2}{lr} & batch & low \\
        & & & size & rank & & size & rank & & size & rank \\
        \hline
        SVDQAT & \WA{4}{8} & $10^{-5}$ & 128 & 32 & $10^{-6}$ & 128 & 16 & $10^{-6}$ & 256 & 16 \\
        vQAT & \WA{4}{8} & $10^{-5}$ & 256 & - & $10^{-6}$ & 128 & - & $10^{-6}$ & 128 & - \\
        \short & \WA{4}{8} & $10^{-6}$ & 256 & - & $10^{-7}$ & 128 & - & $10^{-6}$ & 128 & - \\
    \midrule
    \end{tabular}%
    }
\end{table}

For all \short{} experiments, we follow the schedule highlighted in \tabref{tab:schedule}.

\begin{table}[h]
    \centering
    \caption{\textbf{Precision schedule:} \label{tab:schedule}
    During training we progressively reduce the precision following the prescribed schedule.
    }
    \begin{tabular}{cccccccccc}
         Precision   & $8$ & $7$ & $6$ & $5.5$ & $5$ & $4.75$ & $4.5$ & $4.25$ & $4$ \\
         \hline
         $\#$ epochs & $1$ & $1$ & $1$ & $1$ & $1$ & $2$ & $2$ & $2$ & $14$ \\
    \end{tabular}
    
\end{table}

Experiments with the configuration highlighted above take on average $192$ GPUh for Sana, $576$ GPUh for \pixart, $1008$ GPUh for SD3.5-Medium.

\subsection{Qualitative evaluation}
For additional qualitative evaluation on MJHQ dataset\cite{li2024playground}, please see the \texttt{html} pages in the zip file.

\subsection{Quantitative evaluation}
Here we report additional evaluation of the proposed approach with a wider set of metrics. In particular, we rely on VQA~\cite{lin2024evaluating} to measure the adherence of the generated samples to the input prompts. We measure the image quality with CLIP-IQA~\cite{wang2023exploring}.

\tabref{tab:evaluation_full} shows \short{} outperforms even the strongest QAT baseline we developed namely SVDQAT, with overall higher gains for SD3.5-Medium and \pixart{} for both test datasets.

\begin{table}[t]
    \centering
    \caption{\textbf{Qualitative evaluation:}\label{tab:evaluation_full}
    we evaluate \short{} using a \textit{fractional quantization schedule} on PixArt Evaluation dataset~\cite{chen2024pixart} and  MidJourney HQ Evaluation dataset~\cite{li2024playground} measuring FID, and CLIP-FID wrt the original model, CLIP-IQA~\cite{wang2023exploring}, ImageReward (IR)~\cite{xu2023imagereward}, and VQA~\cite{lin2024evaluating}.}
    \resizebox{\columnwidth}{!}{%
    \begin{tabular}{llcccccccccccccccccccc}
        \toprule
        \cols{22}{\pixart} \\
        \toprule
        & & \cols{5}{SD3.5 Medium} & \cols{5}{Sana 600M} & \multicolumn{5}{c}{\pixart} & \multicolumn{5}{c}{Flux-schnell} \\
        \cmidrule(r){3-7} 
        \cmidrule(r){8-12}
        \cmidrule(r){13-17}
        \cmidrule(r){18-22}
        \rows{2}{Method} & \rows{2}{Precision} & \rows{2}{FID $\downarrow$} & CLIP & CLIP & \rows{2}{IR $\uparrow$} & VQA & \rows{2}{FID $\downarrow$} & CLIP & CLIP & \rows{2}{IR $\uparrow$} & VQA & \rows{2}{FID $\downarrow$} & CLIP & CLIP & \rows{2}{IR $\uparrow$} & VQA & \rows{2}{FID $\downarrow$} & CLIP & CLIP & \rows{2}{IR $\uparrow$} & VQA \\
        & & & FID $\downarrow$ & IQA $\uparrow$ & & score $\uparrow$ & & FID $\downarrow$ & IQA $\uparrow$ & & score $\uparrow$ & & FID $\downarrow$ & IQA $\uparrow$ & & score $\uparrow$ & & FID $\downarrow$ & IQA $\uparrow$ & & score $\uparrow$ \\
        \toprule
        Dynamic Q. & \WA{4}{8} & 9.36 & 2.08 & 0.44 & 0.56 & 0.84 & 2.22 & 0.24 & \textbf{0.46} & 0.57 & \textbf{0.82} & 13.35 & 6.19 & 0.44 & 0.35 & 0.82 & 8.17 & 1.13 & \textbf{0.43} & -0.73 & 0.77 \\
        DiTAS & \WA{4}{8} & 27.93 & 13.77 & \textbf{0.47} & 0.41 & 0.82 & 12.87 & 4.58 & 0.45 & \textbf{0.62} & \textbf{0.82} & 7.30 & 3.95 & \textbf{0.46} & \textbf{0.84} & \textbf{0.86} & - & - & - & - & - \\
        SVDQuant & \WA{4}{16} & 14.42 & 3.14 & 0.42 & 0.66 & 0.85 & 2.43 & 0.24 & 0.43 & 0.60 & \textbf{0.82} & 6.80 & 2.02 & 0.43 & 0.79 & \textbf{0.86} & \textbf{2.26} & 0.36 & 0.42 & 0.84 & 0.85 \\
        \midrule
        SVDQAT & \WA{4}{8} & 2.57 & 0.28 & 0.45 & 0.80 & 0.85 & \textbf{1.93} & \textbf{0.13} & 0.43 & 0.48 & \textbf{0.82} & 5.38 & 1.48 & 0.43 & 0.76 & \textbf{0.86} & - & - & - & - & - \\ 
        vQAT & \WA{4}{8} & 2.67 & 0.31 & 0.44 & 0.78 & 0.85 & 2.13 & 0.16 & 0.43 & 0.45 & 0.81 & 7.00 & 2.52 & 0.45 & 0.79 & 0.85 & 3.40 &0.66 & 0.41 & \textbf{0.87} & \textbf{0.86} \\ 
        \rowcolor{lightblue} \textbf{\short} & \WA{4}{8} & \textbf{2.54} & \textbf{0.27} & 0.45 & \textbf{0.82} & \textbf{0.86} & 2.17 & 0.19 & 0.42 & 0.48 & \textbf{0.82} & \textbf{4.48} & \textbf{1.07} & 0.45 & 0.79 & \textbf{0.86} & 2.55 & \textbf{0.30} & 0.41 & 0.86 & 0.85 \\
        
        \toprule
        \cols{22}{MJHQ} \\
        \toprule
        & & \cols{5}{SD3.5 Medium} & \cols{5}{Sana 600M} & \multicolumn{5}{c}{\pixart} & \multicolumn{5}{c}{Flux-schnell} \\
        \cmidrule(r){3-7} 
        \cmidrule(r){8-12}
        \cmidrule(r){13-17}
        \cmidrule(r){18-22}
        \rows{2}{Method} & \rows{2}{Precision} & \rows{2}{FID $\downarrow$} & CLIP & CLIP & \rows{2}{IR $\uparrow$} & VQA & \rows{2}{FID $\downarrow$} & CLIP & CLIP & \rows{2}{IR $\uparrow$} & VQA & \rows{2}{FID $\downarrow$} & CLIP & CLIP & \rows{2}{IR $\uparrow$} & VQA & \rows{2}{FID $\downarrow$} & CLIP & CLIP & \rows{2}{IR $\uparrow$} & VQA \\
        & & & FID $\downarrow$ & IQA $\uparrow$ & & score $\uparrow$ & & FID $\downarrow$ & IQA $\uparrow$ & & score $\uparrow$ & & FID $\downarrow$ & IQA $\uparrow$ & & score $\uparrow$ & & FID $\downarrow$ & IQA $\uparrow$ & & score $\uparrow$ \\
        \toprule
        Dynamic Q. & \WA{4}{8} & 10.29 & 2.11 & 0.44 & 0.65 & 0.79 & 2.40 & 0.28 & \textbf{0.45} & 0.63 & 0.74 & 15.04 & 5.55 & 0.43 & 0.44 & 0.74 & 8.66 & 1.24 & 0.42 & -0.90 & 0.65 \\
        DiTAS & \WA{4}{8} & 32.04 & 14.06 & \textbf{0.47} & 0.41 & 0.73 & 12.91 & 5.59 & \textbf{0.45} & \textbf{0.68} & \textbf{0.75} & 8.63 & 4.07 & \textbf{0.46} & \textbf{1.04} & 0.80 & - & - & - & - & - \\
        SVDQuant & \WA{4}{16} & 15.10 & 3.06 & 0.42 & 0.78 & 0.78 & 2.48 & 0.25 & 0.42 & 0.62 & \textbf{0.75} & 6.95 & 1.71 & 0.43 & 0.99 & 0.80 & \textbf{2.41} & 0.41 & \textbf{0.42} & 0.96 & 0.79 \\
        \midrule
        SVDQAT & \WA{4}{8} & 2.85 & \textbf{0.32} & 0.45 & 0.91 & 0.80 & \textbf{2.04} & \textbf{0.16} & 0.43 & 0.53 & 0.74 & 5.83 & 1.44 & 0.43 & 0.96 & \textbf{0.81} & - & - & - & - & - \\ 
        vQAT & \WA{4}{8} & 3.01 & 0.37 & 0.44 & 0.89 & 0.80 & 2.13 & 0.20 & 0.43 & 0.47 & 0.74 & 7.38 & 2.12 & 0.44 & 0.99 & 0.80 & 3.56 & 0.73 & 0.41 & \textbf{0.99} & \textbf{0.80} \\ 
        \rowcolor{lightblue} \textbf{\short} & \WA{4}{8} & \textbf{2.78} & \textbf{0.32} & 0.45 & \textbf{0.96} & \textbf{0.81} & 2.34 & 0.24 & 0.42 & 0.50 & 0.74 & \textbf{4.95} & \textbf{1.05} & 0.44 & 0.97 & 0.80 & 2.55 & \textbf{0.39} & 0.41 & \textbf{0.99} & \textbf{0.80} \\
        \midrule
    \end{tabular}%
    }
    \vspace{-0.5cm}
\end{table}

\subsection{Quantization schedule}

\begin{figure}[ht] 
\centering
\includegraphics[width=\linewidth]{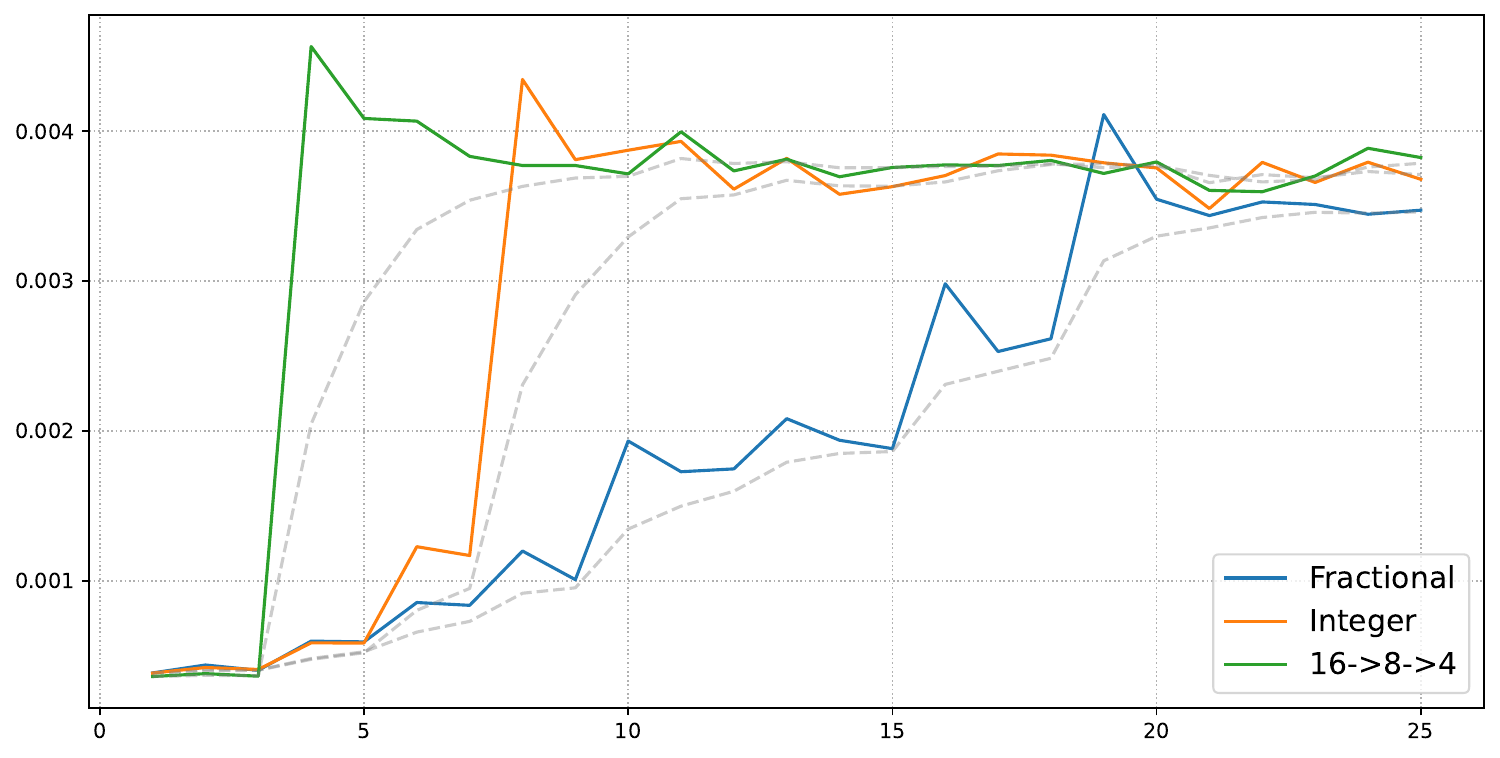}
\caption{\textbf{Fractional schedule:}\label{fig:fractional_val} we train SD3.5-M using a simple progressive schedule (green), an integer schedule (orange), and a fractional schedule (blue). As seen in the graph, the Fractional schedule achives a lower validation loss.}
\end{figure}

To validate the benefits of a \textit{Fractional} quantization schedule (\tabref{tab:schedule}) we compare it with its \textit{Integer} counterpart ($8 \rightarrow 7 \rightarrow 6 \rightarrow 5 \rightarrow \rightarrow 4$), and a simpler progressive schedule ($16 \rightarrow 8 \rightarrow 4$). For a fair comparison, all experiments have the same computational budget. We measure the validation loss across training and plot it in \figref{fig:fractional_val}. The integer and simple schedules perform comparably to each other. On the other hand, the Fractional schedule consistently outperforms the two competitors during training, resulting in a sensibly lower validation loss.

\section{Additional evaluation}

\subsection{Language Model}
The proposed method is agnostic to the architecture and the application. We apply \short{} to Gemma2 2B IT~\cite{gemma_2024}\footnote{\url{https://huggingface.co/google/gemma-2-2b-it}}. Start from the \FP{16} model -- original --, then we quantize it to 4 bits in a similar fashion as we did with T2I models in the main paper. We follow the same schedule as in \secref{sec:params}. The quantized model (\WA{4}{8}) is then compared with the original model. 

As training set we rely on a subset of C4 dataset~\cite{raffel2020exploring}: we pick randomly 384K samples for training and 38.4K samples for validation. The model is evaluated on two datasets in zero-shot fashion: BoolQ~\cite{clark2019boolq}\footnote{\url{https://huggingface.co/datasets/google/boolq}}, and Commonsense QA~\cite{talmor2018commonsenseqa}\footnote{\url{https://huggingface.co/datasets/allenai/social_i_qa}}.
\tabref{tab:eval_llm} shows minimal drop when \short{} is applied to Gemma2 2B IT model. Therefore, proving \short{} can be applied to Language Models as well as Vision Models.

\begin{table}
    \centering
    \caption{\textbf{Evaluation on Language Models:} \label{tab:eval_llm}
    We apply \short{} to Gemma2 2B IT \cite{gemma_2024} exactly as we did to T2I models. The quantized model is evaluated on Common Sense QA and BoolQ datasets. The resulting model has minimal drop from original language model.}
    \begin{tabular}{lcccc}
        \toprule
        Model & Precision & Common Sense QA $\uparrow$ & BoolQ $\uparrow$ & COQA $\uparrow$ \\
        \midrule
        \midrule
        Original & (\WA{16}{32}) & \cfirst $0.70 \pm 0.01$ & \cfirst $0.76 \pm 0.01$ & $0.66 \pm 0.01$ \\
        \short & (\WA{4}{8}) & $0.69 \pm 0.01$ & $0.72 \pm 0.01$ & \cfirst $0.70 \pm 0.01$ \\
        \midrule
    \end{tabular}
\end{table}

\end{document}